\documentclass[letterpaper, 10 pt, conference]{ieeeconf}  

\IEEEoverridecommandlockouts                              

\usepackage{amssymb}
\usepackage{amsmath}
\usepackage{booktabs}
\usepackage{multirow}
\usepackage[dvipdfmx]{graphicx}
\usepackage{subcaption}
\usepackage{wrapfig}
\usepackage{bm}
\usepackage{url}
\usepackage{pifont}
\usepackage{xcolor}
\newcommand{\cmark}{\textcolor{green!80!black}{\ding{51}}}
\newcommand{\xmark}{\textcolor{red}{\ding{55}}}

\title{\LARGE \bf
ScanDP: Generalizable 3D Scanning with Diffusion Policy
}

\author{
    Itsuki Hirako$^{1}$,
    Ryo Hakoda$^{1}$,
    Yubin Liu$^{1}$,
    Matthew Hwang$^{1}$,\\
    Yoshihiro Sato$^{2}$ and
    Takeshi Oishi$^{1}$
\thanks{$^{1}$Itsuki Hirako, Ryo Hakoda, Yubin Liu, Matthew Hwang, and Takeshi Oishi are with Institute of Industrial Science, The University of Tokyo, Japan
        {\tt\small \{hirako, r-hakoda, yubinliu, mhwang, oishi\}@cvl.iis.u-tokyo.ac.jp}}%
\thanks{$^{2}$Yoshihiro Sato is with Kyoto University of Advanced Science, Kyoto, Japan
        {\tt\small sato.yoshihiro@kuas.ac.jp}}%
}

\begin{document}

\maketitle
\thispagestyle{empty}
\pagestyle{empty}

\begin{abstract}

Learning-based 3D Scanning plays a crucial role in enabling efficient and accurate scanning of target objects.
However, recent reinforcement learning-based methods often require large-scale training data and still struggle to generalize to unseen object categories.
In this work, we propose a data-efficient 3D scanning framework that uses Diffusion Policy to imitate human-like scanning strategies.
To enhance robustness and generalization, we adopt the Occupancy Grid Mapping instead of direct point cloud processing, offering improved noise resilience and handling of diverse object geometries.
We also introduce a hybrid approach combining a sphere-based space representation with a path optimization procedure that ensures path safety and scanning efficiency. 
This approach addresses limitations in conventional imitation learning, such as redundant or unpredictable behavior.
We evaluate our method on diverse unseen objects in both shape and scale.
Ours achieves higher coverage and shorter paths than baselines, while remaining robust to sensor noise.
We further confirm practical feasibility and stable operation in real-world execution.
\end{abstract}

\section{INTRODUCTION}

3D scanning plays an essential role across a wide range of fields, including robotics~\cite{shen2023distilled}, autonomous driving\cite{9455394}, industrial inspection~\cite {haleem2022exploring}, and digital archival~\cite{ikeuchi2007great,nemoto2023virtual}.
Typical examples of 3D scanning systems include handheld scanners operated by humans and sensor-equipped platforms such as drones or mobile robots that are actively controlled to acquire data.
However, human-operated 3D scanning faces significant challenges, such as substantial time consumption, labor intensiveness, and vulnerability to human errors.
Obtaining complete surface data can require several hours even for objects comparable in size to a statue, while scanning large structures like buildings may take several days.
Moreover, errors during the scanning process may necessitate re-scanning, further impeding the efficiency of high-quality 3D data acquisition.

In light of these challenges, recent research has actively explored the automation of 3D scanning.
Automated 3D scanning approaches can generally be categorized into rule-based and learning-based methods.
Rule-based methods efficiently determine scanning positions according to predefined strategies.
A common example is mounting a scanner on the end-effector of a robotic arm to scan objects placed on a turntable~\cite{border2024surface}.
For large-scale environments, autonomous drone-based 3D mapping systems equipped with LiDAR sensors have also been developed~\cite{ren2025safety,tang2023bubble}.
In contrast, learning-based methods seek to optimize scanning strategies through trial and error through reinforcement learning (RL)~\cite{chen2024gennbv, peralta2020next}.
Although learning-based approaches have demonstrated strong performance for specific categories of objects, challenges remain in enhancing generalizability and reducing computational costs.


\begin{figure}
    \centering
    \begin{subfigure}{0.95\linewidth}
        \centering
        \includegraphics[width=\linewidth]{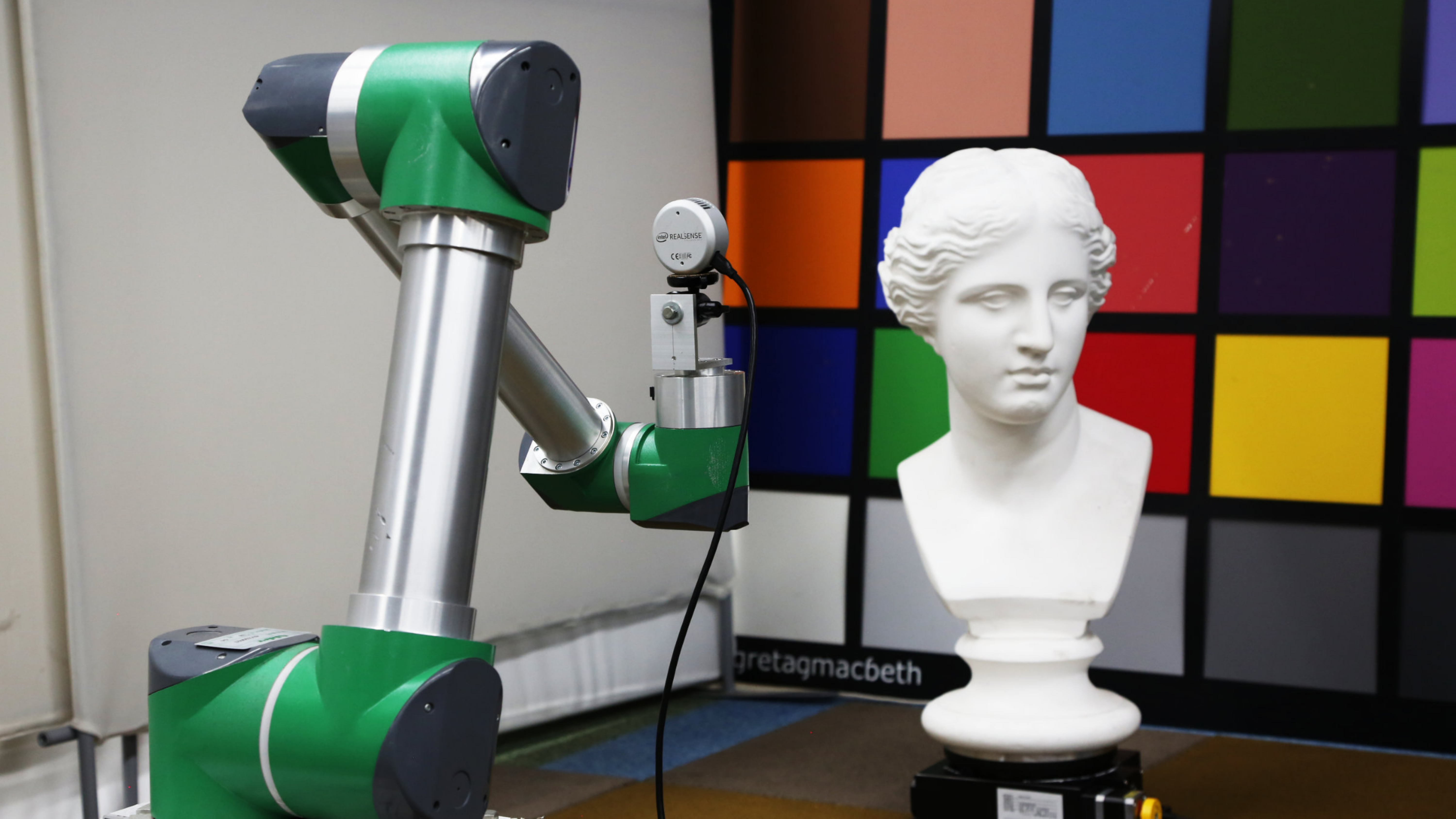}
        \vspace{1 mm}
    \end{subfigure}
    \begin{subfigure}{\linewidth}
        \centering
        \includegraphics[width=\linewidth]{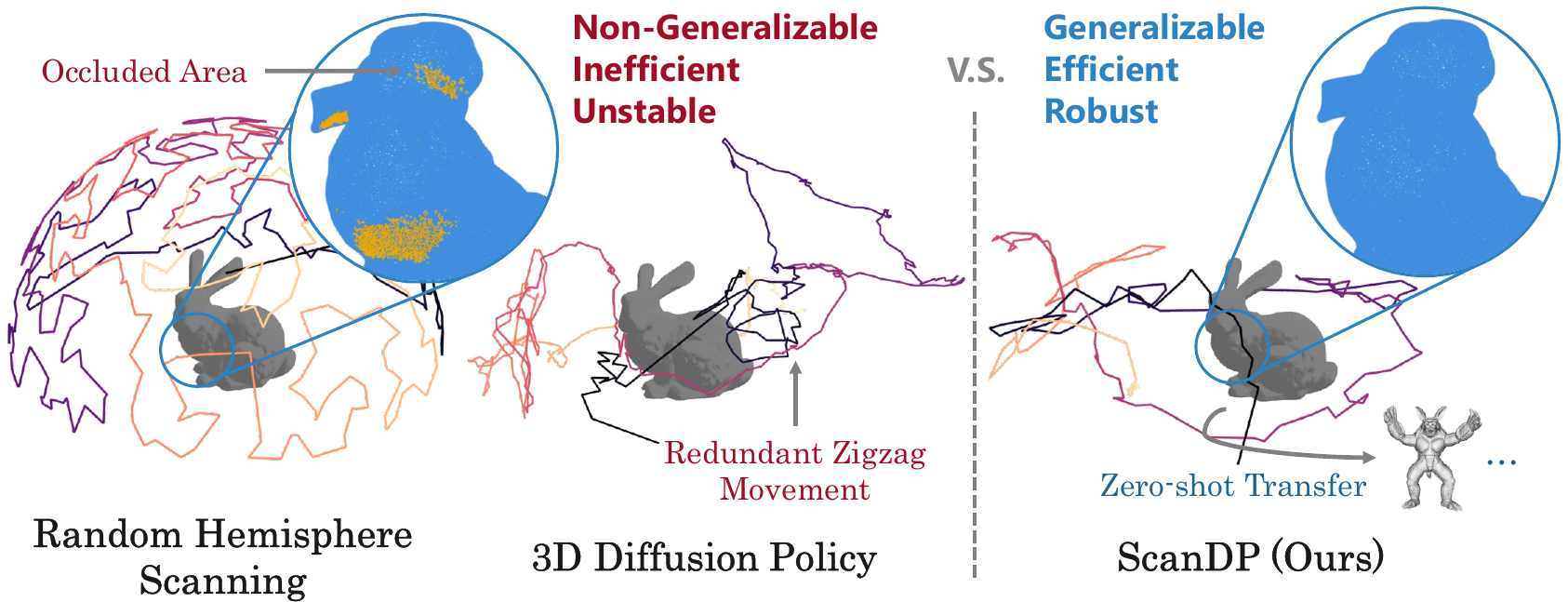}
    \end{subfigure}
      \caption{\textbf{Generalizable 3D Scanning Policy.}
      Our proposed framework ScanDP can generalize to unseen objects with a small amount of training data (Right).
      Prior works lack generalization capability and are not robust to changes in conditions (left).
      }
      \label{fig:teaser}
\end{figure}

To address these challenges, imitation learning (IL) has emerged as a promising solution.
IL enables the acquisition of high-precision policies with high sample efficiency by directly imitating expert demonstrations~\cite{chi2025diffusion,florence2022implicit,zhao2023learning,ma2024hierarchical}.
In particular, for complex 3D scanning tasks where explicit reward design for RL is difficult, IL provides a data-driven approach to acquiring effective strategies.
Recently, Diffusion Policy~\cite{chi2025diffusion}, a framework that applies conditional diffusion models to continuous control problems, has demonstrated high flexibility and powerful modeling capabilities.
However, IL still faces challenges, including the occurrence of unexpected behaviors~\cite{ze20243d} and the generation of suboptimal actions.

Therefore, this paper proposes a method for autonomous 3D scanning that achieves high generalization performance and optimal path planning.
The proposed method is built upon Diffusion Policy, a framework within IL.
While existing approaches typically rely on point clouds to perceive the 3D environment, our method utilizes an occupancy grid map (OGM)~\cite{thrun2002probabilistic}, enabling not only spatial understanding but also integrated recognition of measurement uncertainties.
Furthermore, to mitigate the issues of unexpected behaviors and suboptimal actions, we introduce a maximum empty sphere (bubble) representation combined with path optimization techniques to enhance robustness and overall performance.

The main contributions of this work are summarized as follows: 
\begin{enumerate}
    \item \textbf{Generalizability.}The proposed method achieves high generalization performance even on unseen and completely different types of objects, demonstrating strong adaptability across diverse scanning targets. 
    \item \textbf{Efficiency.} It enables the training of high-performing models with a limited amount of expert data, reducing the burden of extensive data collection efforts. 
    \item \textbf{Robustness.} The method shows robustness to robot motion disturbances and noisy input conditions, suggesting its potential for stable operation in practical scenarios. 
\end{enumerate}

We evaluate our method across various object categories, measuring coverage and path length.
Our approach consistently achieves higher coverage than baselines, which often become stuck at certain viewpoints.

\begin{table}[t]
  \centering
  \scriptsize
  \caption{\textbf{Comparison of prior works.}}
  \label{tab:works}
  \resizebox{0.99\linewidth}{!}{
    \begin{tabular}{lcccccr}
      \toprule
      \multirow{2}{*}{} & \multicolumn{3}{c}{Input} & \multicolumn{2}{c}{Camera} & \multirow{2}{*}{Encoder} \\
      \cmidrule(lr){2-4}\cmidrule(lr){5-6}
      & Pose & Image & Point Cloud & Count & Mobile & \\
      \midrule
      Diffusion Policy~\cite{chi2025diffusion} & \cmark & \cmark & \xmark & 2 & \xmark & ResNet \\
      3D Diffusion Policy~\cite{ze20243d} & \cmark & \xmark & \cmark & 1 & \xmark & MLP \\
      GenDP~\cite{wang2024gendp} & \cmark & \cmark & \cmark & 4 & \xmark & PointNet++\cite{qi2017pointnet++} \\
      \midrule
      \textbf{ScanDP (ours)} & \cmark & \xmark & \cmark & 1 & \cmark & SparseConv\cite{choy20194d} \\
      \midrule
    \end{tabular}
  }
\end{table}

\section{Related Work} \label{sec:related_work}

\subsection{3D Scanning}
\textbf{Rule-based methods.} An illustrative example of a rule-based approach is frontier-based exploration (FBE)~\cite{yamauchi1997frontier, tang2023bubble}, which uses heuristics to select the optimal frontier at the boundary of known regions when deciding the next action.
However, due to their heuristic nature, rule-based methods cannot adapt based on prior data and tend to perform poorly in complex environments.
To address these limitations, learning-based methods have been actively studied in recent years.

\textbf{Learning-based methods.} Learning-based approaches primarily utilize reinforcement learning and are often framed as a Next-Best View (NBV) problem.
Some studies optimize viewpoints from a set of predefined views on a hemisphere~\cite{zhan2022activermap, 9913658}, while others optimize viewpoints freely in space~\cite{peralta2020next, zeng2020pc}.
However, these methods typically require training on hundreds of objects to achieve good generalization, leading to high training costs.
Moreover, RL approaches require extensive reward design, which can be challenging and labor-intensive.

Hence, in this work, we propose a 3D Scanning method based on IL, which offers higher training efficiency and better generalization capabilities.

\subsection{Diffusion Models for Robotics}
Recently, approaches incorporating Diffusion Models into IL have attracted significant attention.
Diffusion Models, originally developed in the field of image generation, have demonstrated strong expressiveness across a variety of generative tasks~\cite{ho2020denoising, song2020denoising}.
Diffusion Models learn data distributions by progressively adding noise to data and subsequently learning to denoise through a neural network.
One of the key strengths of Diffusion Models is their ability to model high-dimensional continuous distributions, enabling stable learning.
Moreover, Diffusion Models can generate conditioned outputs by incorporating modalities such as text and images.

According to these characteristics, the application of Diffusion Models to robotics has recently been extensively explored, beginning with the introduction of the Diffusion Policy (DP) in manipulation~\cite{chi2025diffusion, ze20243d}, followed by navigation~\cite{sridhar2024nomad}, planning~\cite{cao2024dare}, and integration with large language models (LLMs)~\cite{ha2023scaling}.
The aforementioned DP treats images as observations and employs a Diffusion Model to generate actions in the context of imitation learning.
However, since DP relies on RGB images, it struggles to capture the three-dimensional spatial structure of environments, resulting in limited robustness to variations in camera viewpoints, surrounding environments, and object properties.

3D Diffusion Policy (DP3) addresses this limitation by utilizing point clouds instead of images as observations, resulting in improved generalization across objects of different sizes and morphology~\cite{ze20243d}.

However, little research has focused on leveraging Diffusion Policy in the context of 3D scanning.
Furthermore, given that the objects subjected to scanning may be of significant cultural value, the risk of collision must be eliminated, necessitating safe path optimization.

To generate optimal long-horizon actions, we adopt the OGM for the observation input instead of point clouds, enabling an explicit representation of the whole scanning process.
In our experiments, we set DP and DP3 as baselines for comparison and evaluation.
Please refer to Tab.~\ref{tab:works} for the differences in conditions between the proposed method and existing approaches.

\begin{figure*}[t]
  \centering
  \includegraphics[width=1\linewidth]{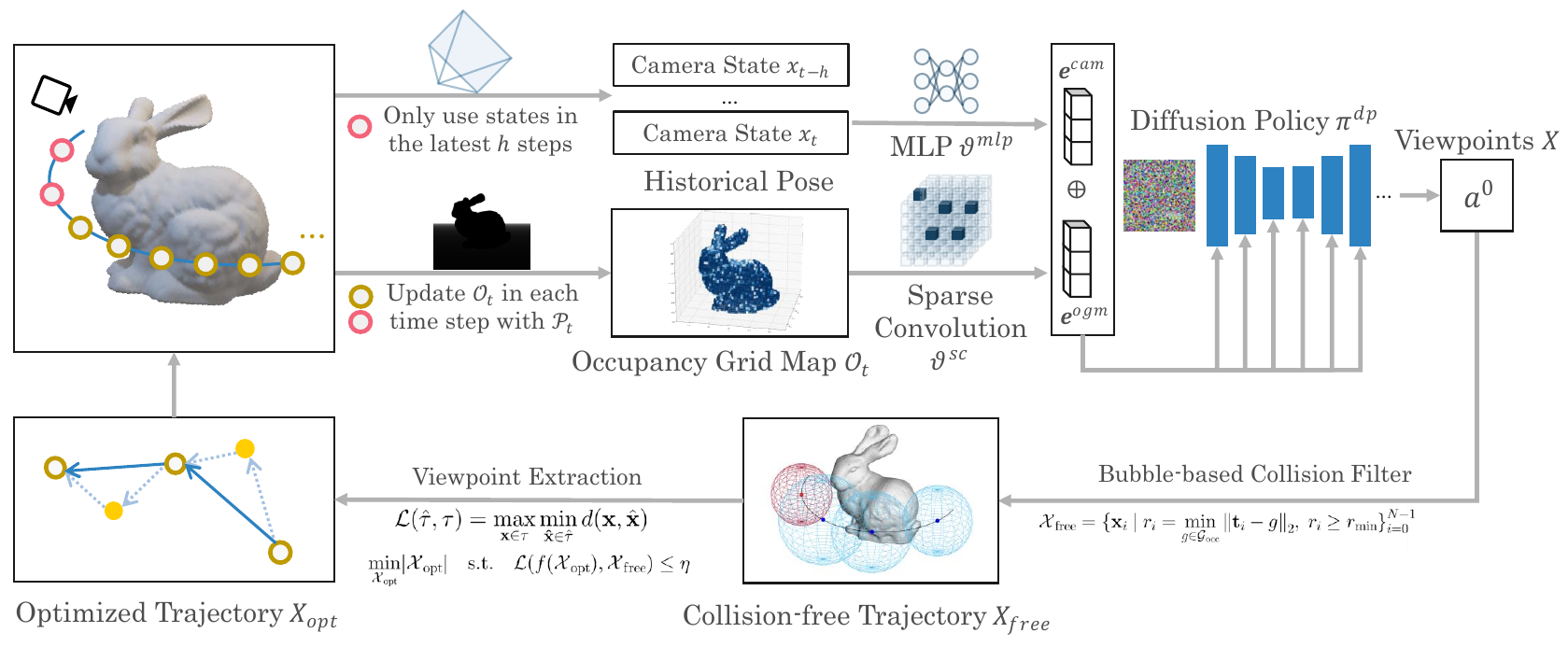}
  \caption{\textbf{Method Overview.}
  ScanDP consists of two main components: Path Generation (Upper part) and Path Optimization (Lower part).
  In the Path Generation phase, actions are generated from a diffusion policy conditioned on an occupancy grid map (OGM) and the current camera pose.
  In the Path Optimization phase, the generated actions are refined through a bubble-based collision filter and viewpoint extraction to optimize the scanning trajectory.}
  \label{fig:overview}
\end{figure*}

\section{Methodology}
\label{sec:methology}

\subsection{Overview}
\label{subsec:problem_definition}

We design \textbf{ScanDP} to perform efficient and high-precision 3D scanning by generating a global path for camera movement. 
At each time step $t \in \mathbb{N}$, ScanDP receives a camera pose $\bm{x}_t \in \mathrm{SE}(3)$ and a depth map $\mathcal{D}_t \in \mathbb{R}^{\mathrm{H} \times \mathrm{W}}$ as the observation $\bm{o}_t = \{\bm{x}_t, \mathcal{D}_t\}$. 
$\mathrm{H} \times \mathrm{W}$ is the resolution of the input depth map. 
In this work, we treat the pose $\bm{x}$ as the action space. 
The policy takes the past $h$ observations $\bm{o}_{t-h+1:t}$ as input and outputs the next $N$ actions $\bm{a}_{t:t+N-1}$. 
ScanDP continues to move until it reaches a predefined time limit $T$. 
The final output is a point cloud constructed from all depth maps $\mathcal{D}_{1:T}$. 

ScanDP is a visuomotor policy that achieves precise and efficient path generation by learning from a small number of human scanning demonstrations. 
An overview of the framework is shown in Fig.~\ref{fig:overview}. 
ScanDP consists of two main parts: 
(a)~\textbf{Path Generation}, where actions are generated from a diffusion policy conditioned on depth images and previous camera poses, and 
(b)~\textbf{Path Optimization}, where the generated actions are refined using a minimal sphere representation and dynamic programming.  
The following sections describe each component in detail.

\subsection{Path Generation}
\textbf{OGM Representation.}
At time $t$ OGM is defined by a 3D voxel grid $\mathcal{M} = \{\bm{m}\}$ and their occupancy probabilities $p$: $\mathcal{O}_t = \{ (\bm{m}, p(\bm{m}|\bm{o}_{1:t}))|\bm{m} \in \mathcal{M} \}$.
OGM integrates measurements up to time step $t$ based on a Bayesian update, using the camera pose $\bm{x}_t$ and the corresponding point cloud $\mathcal{P}_t$ as inputs at the $t$-th update~\cite{thrun2002probabilistic}.
From the depth map $\mathcal{D}_t$ and camera pose $\bm{x}_t$, we obtain a point cloud $\mathcal{P}_t$ in the world coordinate system: $\mathcal{P}_t = \bm{x}_t \circ \pi^{-1}(\mathcal{D}_t)$. 
Here $\pi$ is the projection function, and $\circ$ is the rigid transformation. 
Bresenham's line algorithm~\cite{5388473} is used to determine the grid cells to be updated: the grid containing $\mathcal{P}_t$ is designated as the $\bm{l}_\mathrm{hit}$, and the grids along the line segment between the camera center and $\mathcal{P}_t$ are designated as $\bm{l}_\mathrm{miss}$. 

Assuming that the OGM update follows a Markov process, we incrementally update the occupancy probability $p(\bm{m}|\bm{x}_{1:t}, \mathcal{P}_{1:t})$ of a grid $\bm{m}$ using the log-odds belief~\cite{thrun2002probabilistic,hornung2013octomap}. 
The odds ratio of the occupancy probability is derived as follows:

\begin{equation}
  \label{eq:prob}
  \begin{aligned}
  \log &\frac{p(\bm{m}|\bm{x}_{1:t}, \mathcal{P}_{1:t})}{1 - p(\bm{m}|\bm{x}_{1:t}, \mathcal{P}_{1:t})}
  = \log \frac{p(\bm{m}|\bm{x}_{t}, \mathcal{P}_{t})}{1 - p(\bm{m}|\bm{x}_{t}, \mathcal{P}_{t})} \\
  &+ \log \frac{p(\bm{m}|\bm{x}_{1:t-1}, \mathcal{P}_{1:t-1})}{1 - p(\bm{m}|\bm{x}_{1:t-1}, \mathcal{P}_{1:t-1})}
  + \log \frac{1 - p(\bm{m})}{p(\bm{m})}
  \end{aligned}
\end{equation}

By assuming the prior $p(\bm{m})=0.5$ and using the log-odds notation 
$L_{(\cdot)}(\bm{m})= \log\left( \frac{p(\bm{m}|{}_{(\cdot)})}{1-p(\bm{m}|{}_{(\cdot)})}\right)$, 
%
the OGM update process can be performed as follows:
\begin{equation}
  \label{eq:log_odds}
  L_{1:t}(\bm{m}) = L_{1:t-1}(\bm{m}) + L_t(\bm{m}). 
\end{equation}
Following~\cite{hornung2013octomap}, we set $L_t(\bm{m})$ to $0.85$ for $\bm{l}_\text{hit}$ and $-0.4$ for $\bm{l}_\text{miss}$.

In previous studies using OGM \cite{chen2024gennbv, thrun2002probabilistic, hornung2013octomap}, grid cells are often classified into \textit{Free}, \textit{Occupied}, and \textit{Unknown} states based on thresholding.
However, in this work, since we aim to capture the change in $p$, we use the raw probability values without classification during the encoding process.

\textbf{Encoding OGM.} 
To extract features from the OGM, we apply Sparse Convolution~\cite{graham2017submanifold, choy20194d}.
Although 3D Convolution is commonly used for feature extraction from voxel grids, it is inefficient and less effective for sparse tensors, such as OGMs, where many voxels represent free space.
Thus, inspired by prior work~\cite{hoeller2022neural}, we also adopt sparse convolution for feature extraction from the OGM.
The encoder takes the occupancy probabilities of each grid as input and outputs feature representations.
The encoder architecture consists of three sparse convolution layers, an average pooling layer, and a final fully connected layer.
Applying the sparse convolution encoder $\bm{\Theta}^{\mathrm{sc}}$ to the occupancy grid map $\mathcal{O}_t$, we obtain the feature vector:
$\bm{e}^{\text{ogm}} = \bm{\Theta}^{\mathrm{sc}}(\mathcal{O}_t)$. 

\textbf{Action Generation.}
Our policy is modeled as a Denoising Diffusion Probabilistic Model (DDPM)~\cite{ho2020denoising}.
Our model generates actions conditioned on the concatenated features of the OGM feature and the camera pose feature: $\bm{e} = \bm{e}^{\text{cam}} \oplus \bm{e}^{\text{ogm}}$. 
The camera pose $\bm{x}_t$  is encoded using a simple MLP $\bm{\Theta}^{\mathrm{mlp}}$ by the following formula   $\bm{e}^{\text{cam}} = \bm{\Theta}^{\mathrm{mlp}}(\bm{x}_{t-h+1:t})$, where h is the historical observation length.

At inference time, an initial noisy action $\bm{a}^k$ is sampled from a standard normal distribution, and denoising is performed progressively according to the update rule in Eq.~\ref{eq:act}, yielding the final action $\bm{a}^0$.
\begin{equation}
  \label{eq:act}
  \bm{a}^{k-1}=\alpha_k\left(a^k-\gamma_k \boldsymbol{\varepsilon}_\theta\left(\bm{a}^k, k, \bm{e}\right)\right)+\sigma_k \mathcal{N}(0, \mathbf{I}).
\end{equation}
Here, $\bm{\varepsilon}_\theta$ denotes the noise prediction network, which takes as input the noisy action $\bm{a}^k$, timestep index $k$, and conditioned feature $\bm{e}$, and outputs the estimated noise.
$\alpha_k$, $\gamma_k$, and $\sigma_k$ are predefined hyperparameters.

During training, we add randomly sampled noise $\boldsymbol{\varepsilon}^k$ to the noise-free action $\bm{a}^0$ at a randomly selected timestep $k$, and train the network to predict the added noise from the noisy action.
The objective is to minimize the mean squared error (MSE) between the true noise $\bm{\varepsilon}^k$ and the predicted noise, as formulated by the following loss function:
\begin{equation}
  \label{eq:loss}
  \mathcal{L} = \text{MSE}\left(\boldsymbol{\varepsilon}^k, \boldsymbol{\varepsilon}_\theta(\bar{\alpha}_k \bm{a}^0 + \bar{\beta}_k \boldsymbol{\varepsilon}^k, k, \bm{e})\right).
\end{equation}
Through this training process, the network learns to effectively remove noise at each step, allowing it to progressively recover high-quality actions from pure noise during inference.
The output consists of $N$ steps of actions, with the final action denoted as $\bm{a}_{t+N-1}$, and the resulting camera pose horizon is given as:
\begin{equation}
\mathcal{X} = \{\bm{x}_i = \bm{a}_{t+i}\}_{i=0}^{N-1}. 
\end{equation}

\subsection{Path Optimization}
In past research, DP and other IL methods have been reported to behave unexpectedly during inference.
Furthermore, since ScanDP mimics non-optimized human motion, we perform bubble-based collision filtering and viewpoint extraction to guarantee (a)~collision-free and (b)~smooth trajectory.
In practice, we perform the path optimization discussed 
 above on the 16 action steps produced by the Diffusion process. 

\textbf{Bubble-based Collision filter.}
ScanDP verifies the absence of obstacles around the camera using an Occupancy Grid Map (OGM).  
In the OGM, if the occupancy probability $p$ of a grid exceeds a predefined threshold, the grid is considered \textit{Occupied} and is treated as an obstacle.  
Various methods exist for identifying obstacle-free spaces, such as using spheres~\cite{xie2022fast,ren2022bubble,tang2023bubble} or polyhedra~\cite{wang2022geometrically,ren2025safety}.  
In this study, since it is assumed that there are no obstacles near the target object, we adopt the simplest form:a bubble.  
Searching all grids in the OGM is inefficient. 
Therefore, we only consider grids with an occupancy probability greater than or equal to the threshold $\kappa_\text{occ}$ as \textit{Occupied}, and limit the search to those grids. 
For each step's camera translation $\mathbf{t}_i$ included in a camera pose $\bm{x}_i$, we consider a bubble centered at $\mathbf{t}_i$ with its radius $r_i$ defined as the Euclidean distance to the nearest \textit{Occupied} grid.  
Only viewpoints with $r_i \geq r_{\text{min}}$ are considered safe and included in the trajectory.  
Based on previous work~\cite{hornung2013octomap}, we set $\kappa_\text{occ} = 0.9$ and $r_{\text{min}} = 0.1\,\mathrm{m}$.  
The occlusion-free camera pose horizon $\mathcal{X}_\mathrm{free}$ is then defined as:
\begin{equation}
  \label{eq:bubble}
  \mathcal{X}_\text{free} = \{ \bm{x}_i \mid r_i = \min_{g \in \mathcal{G}_{\text{occ}}} \|\mathbf{t}_i - g\|_2,\; r_i \geq r_{\text{min}} \}_{i = 0}^{N-1},
\end{equation}
where $\mathcal{G}_{\text{occ}}=\{g\}$ denotes the set of grid centers marked as \textit{Occupied} in the OGM, respectively.

\begin{table*}[t]
  \centering
  \caption{\textbf{Evaluation results} of Scanning policies.
    All policies are trained only on trajectories scanning the \textbf{Stanford Bunny}. Coverage is shown in mean $\pm$ std.
  We find that ScanDP consistently achieves the highest coverage while maintaining low variance.}
  \label{tab:coverage}
  \resizebox{\linewidth}{!}{
  \begin{tabular}{lll|llllll}
    \toprule
    \multicolumn{1}{l}{Policy} 
      & \multicolumn{2}{c|}{
          \begin{tabular}{c}
            \includegraphics[width=0.1\linewidth]{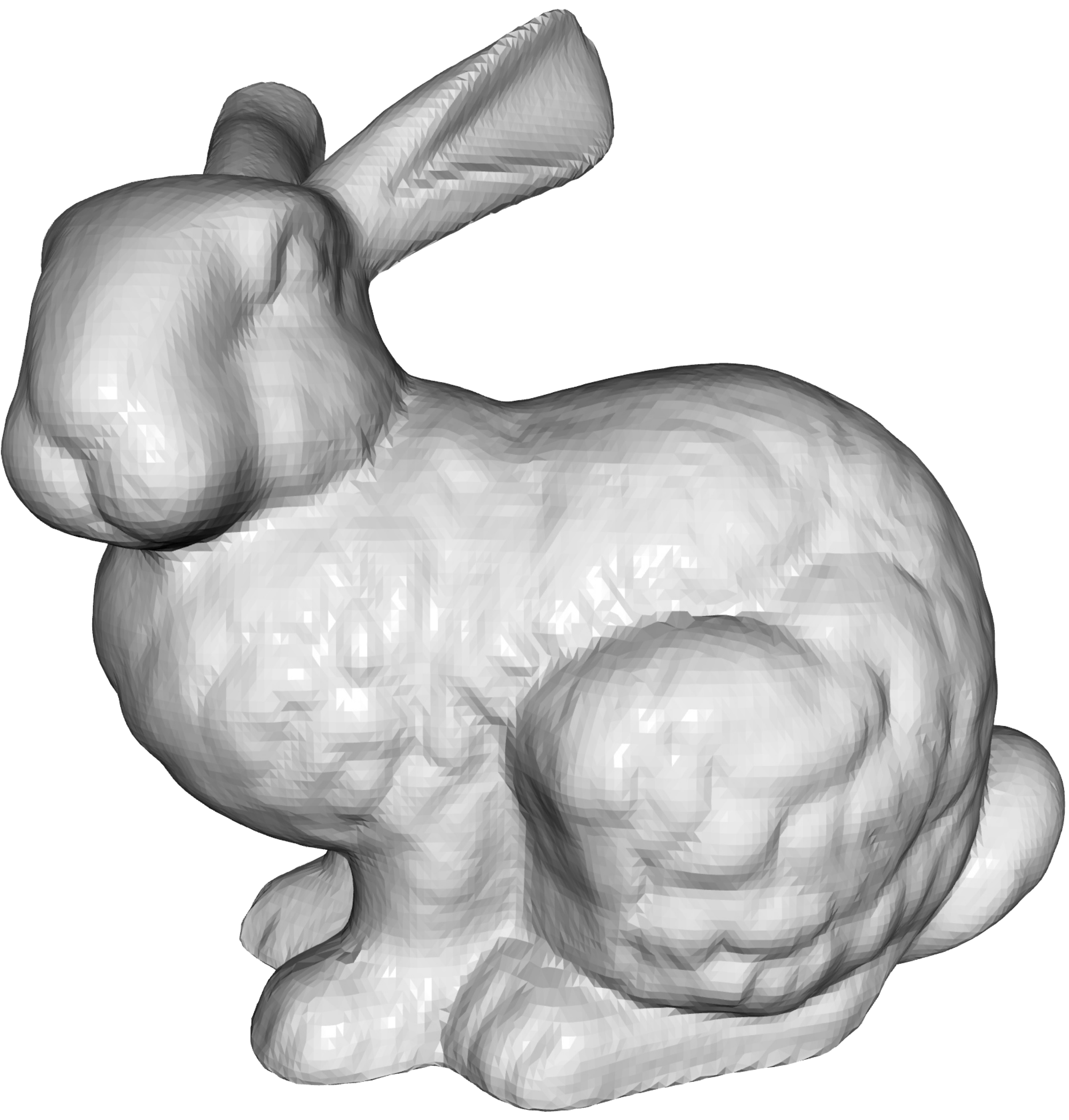} \\
            Bunny
          \end{tabular}
        } 
      & \multicolumn{2}{c}{
          \begin{tabular}{c}
            \includegraphics[width=0.1\linewidth]{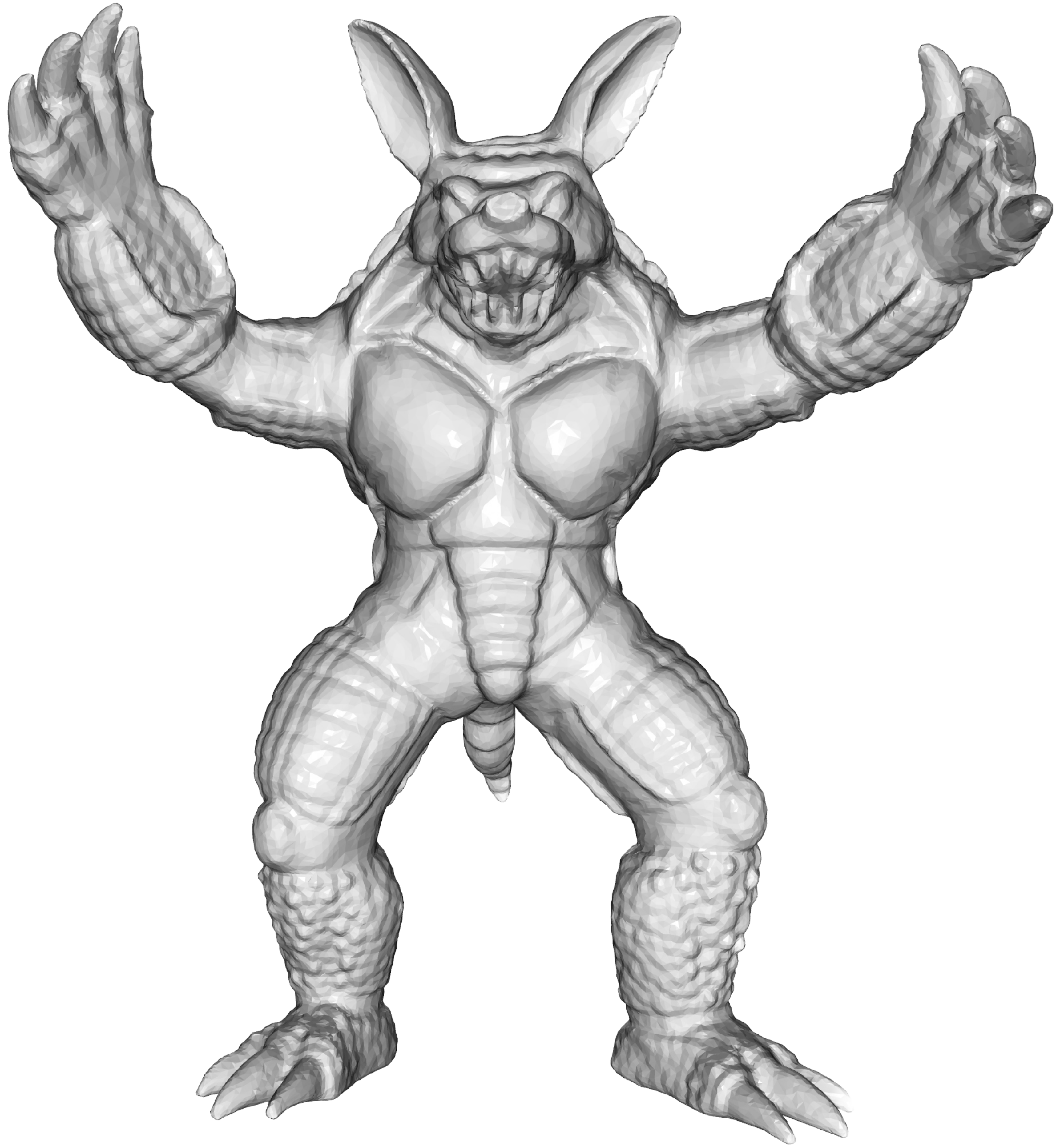} \\
            Armadillo
          \end{tabular}
        } 
      & \multicolumn{2}{c}{
          \begin{tabular}{c}
            \includegraphics[width=0.1\linewidth]{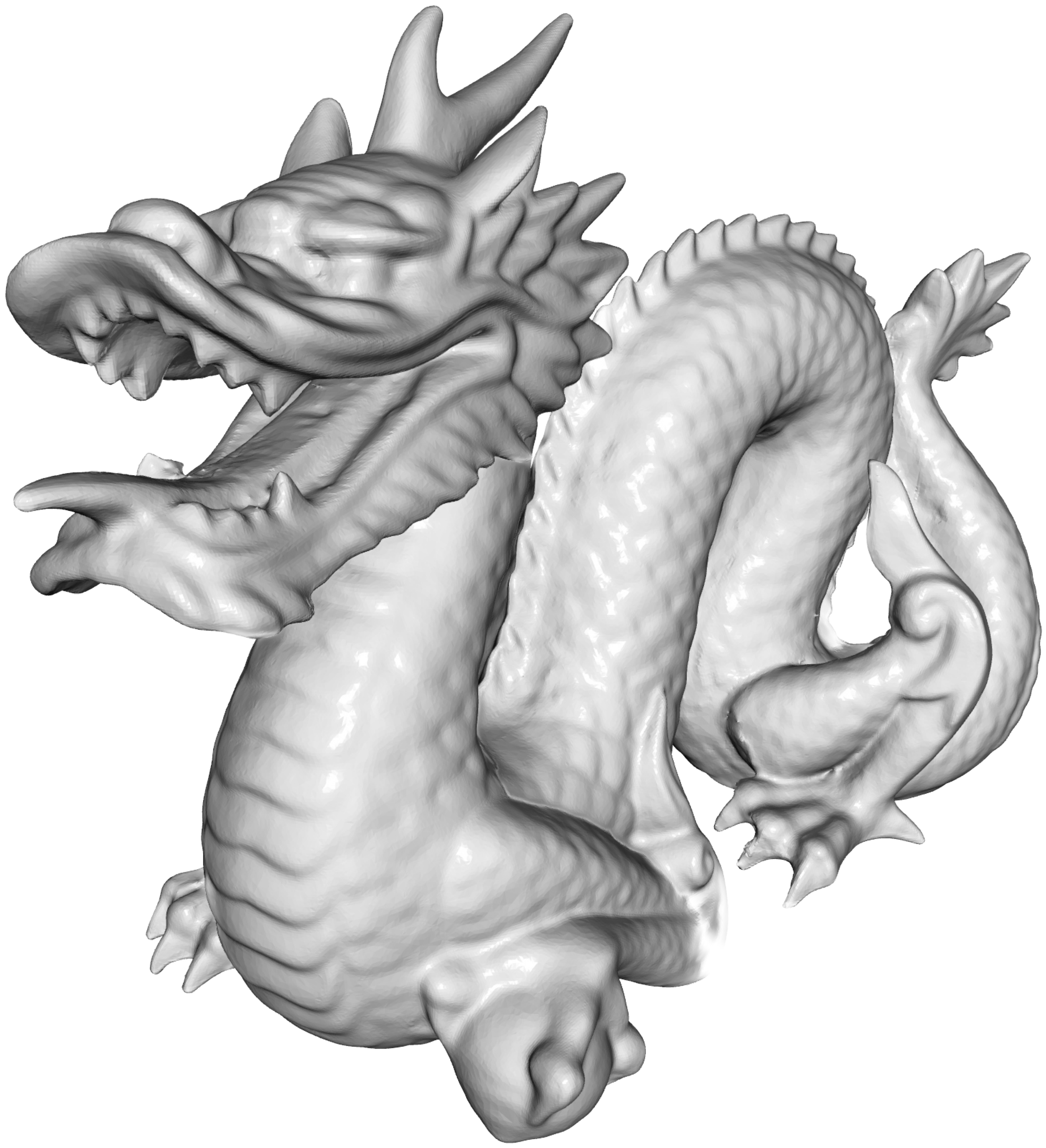} \\
            Dragon
          \end{tabular}
        } 
      & \multicolumn{2}{c}{
          \begin{tabular}{c}
            \includegraphics[width=0.08\linewidth]{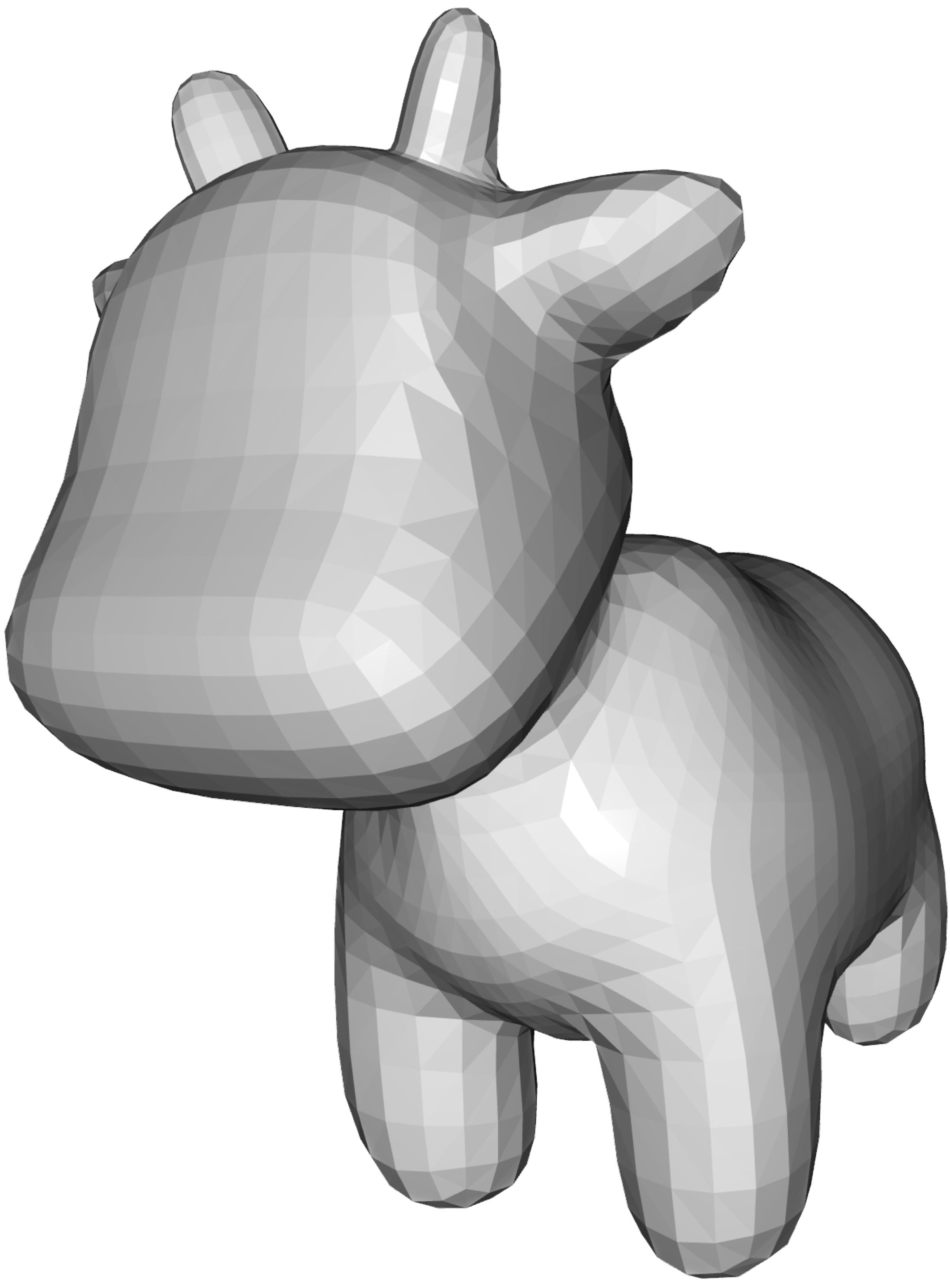} \\
            Spot
          \end{tabular}
        } \\
    \cmidrule(lr){2-3} \cmidrule(lr){4-5} \cmidrule(lr){6-7} \cmidrule(lr){8-9}
      & $\times$1.0 & $\times$1.5 & $\times$1.0 & $\times$1.5 & $\times$1.0 & $\times$1.5 & $\times$1.0 & $\times$1.5 \\
    \midrule
    Random                    & 92.30 & --    & 89.05 & --    & 90.20 & --    & 86.04 & --     \\
    Random Hemisphere         & 85.88 & --    & 95.85 & --    & 88.96 & --    & 96.52 & --    \\
    Uniform Hemisphere        & 85.92 & --    & 96.18 & --    & 88.92 & --    & 86.41 & --    \\
    \midrule
    Diffusion Policy\cite{chi2025diffusion}    &94$\pm$8.9  &73$\pm$9.4    &95$\pm$2.0  &92$\pm$0.8    &89$\pm$1.0 &73$\pm$17.5    &95$\pm$4.8 &78$\pm$1.2    \\
    3D Diffusion Policy\cite{ze20243d}& 94$\pm$9.0 & 79$\pm$17.9 & 94$\pm$8.2 & 88$\pm$8.5 & 93$\pm$4.2 & \textbf{90$\pm$5.9} & 94$\pm$5.6 & 73$\pm$19.6 \\
    \midrule
    \textbf{ScanDP (ours)}     & \textbf{97$\pm$2.4} & \textbf{87$\pm$1.1} & \textbf{99$\pm$0.6} & \textbf{96$\pm$0.9} & \textbf{97$\pm$0.6} & 87$\pm$0.7 & \textbf{97$\pm$2.1} & \textbf{91$\pm$0.2} \\
    \bottomrule
  \end{tabular}
  }
\end{table*}

\textbf{Viewpoint Extraction.} 
Following the pre-processing method used in a previous study \cite{shi2023waypoint}, our method performs the following procedure.
The final output is denoted as $\mathcal{X}_\mathrm{opt}$, and prior to optimization, $\mathcal{X}_\text{opt}$ is initialized to $\mathcal{X}_\text{free}$.

Given a camera pose horizon $\mathcal{X}_\text{free}$ that is guaranteed to be safe by the Bubble method. 
A linear interpolation function $f$ is used to construct the trajectory from the camera pose horizon as: $\tau_\textrm{opt} = f(\mathcal{X}_\text{opt})$.
To evaluate how well $\tau_\mathrm{opt}$ approximates the original path, we define the reconstruction loss $\mathcal{L}$ as follows:
\begin{equation}
  \label{eq:loss_view}
  \mathcal{L}(\hat{\tau}, \tau) = \max_{\bm{x} \in \tau} \min_{\bm{\hat{x}} \in \hat{\tau}} d(\bm{x}, \hat{\bm{x}}),
\end{equation}
where $d(\cdot, \cdot)$ denotes the Euclidean distance function. For each pose $\mathbf{x}_\text{free}\in\mathcal{X}_{free}$, the distance to the nearest point on the approximate trajectory is computed, and the maximum of these distances is taken as the loss.
Based on Eq. \ref{eq:loss_view}, we consider the following optimization problem:
\begin{equation}
  \min_{\mathcal{X}_\textrm{opt}} \left|\mathcal{X}_\textrm{opt}\right| \quad \text{s.t.} \quad \mathcal{L}(f(\mathcal{X}_\mathrm{opt}), \mathcal{X}_\mathrm{free}) \leq \eta.
\end{equation}
The goal is to minimize the number of poses included in $\mathcal{X}_\mathrm{opt}$ while ensuring that the reconstruction loss $\mathcal{L}$ remains below a threshold $\eta$.
This optimization problem is solved using dynamic programming.
Based on the optimized camera pose horizon $\mathcal{X}_\mathrm{opt}$, the camera is moved accordingly.

\section{Simulation Experiments}
\label{sec:experiments}

\subsection{Setup}
\textbf{Simulation Environment.}
All experiments were conducted using Genesis~\cite{Genesis}, a physics simulator for robot learning.
The experiments were run on a system equipped with an NVIDIA GeForce RTX 3090 GPU and an AMD Ryzen 9 5950X 16-core CPU.
The resolution of the acquired $\mathcal{D}_t$ was set to $224 \times 224$ pixels, with a field of view (FoV) of $45^\circ \times 45^\circ$.
The objects to be scanned were prepared to fit within a cubic area with a side length of $0.25~m$.
The OGM covered a cubic area with a side length of $0.8~m$ with the grid size set to $0.02~m$.
In our experiment, action horizon N=16 and observation horizon h=2.

We used the following methods for comparison:
\begin{enumerate}
\item \textbf{Random}: Random actions (camera poses) were generated within the action space, and the shortest path visiting all points was derived using the Traveling Salesman Problem (TSP)~\cite{perron_et_al:LIPIcs.CP.2023.3}.
\item \textbf{Random Hemisphere}: Random points were scattered on the surface of a hemisphere centered on the object, and the shortest TSP path was used to visit all points.
\item \textbf{Uniform Hemisphere}: Uniformly distributed points were generated on the same hemisphere using the Fibonacci lattice, and the shortest TSP path was used to visit all points.
\item \textbf{Diffusion Policy(DP)~\cite{chi2025diffusion}}: A method using the Diffusion Policy. For a fair comparison, we use a single depth image for the input.
\item \textbf{3D Diffusion Policy(DP3)~\cite{ze20243d}}: A variant of the Diffusion Policy which utilizes point clouds. The input is a single depth image, and the camera is mobile rather than being fixed.
\end{enumerate}
Evaluation was performed by running 500 inference steps for each method on each target object.

\textbf{Dataset.}
We used only the Stanford Bunny model from the Stanford 3D Scanning Repository to obtain training data. 
With the model positioned in a simulator environment, we created scanning trajectories by controlling a simulated RGB-D camera using the IMU in an Intel RealSense T265, which is connected to the virtual camera. 
Each scan consists of 500 steps, and the training dataset has five trajectories in total.
In addition to the Stanford 3D Scanning Repository, the evaluation dataset included objects with significant occluded areas, such as the Spot model\cite{crane2013robust}. 
During evaluation, we recorded 500 steps, following the same protocol as in the dataset generation. 
When replaying the demonstration on novel objects (Scale$\times$1.0), the average coverage was 85\%, with self-occluded regions remaining unscanned.

\textbf{Evaluation Metrics.}
We used the coverage metric $C(\mathcal{P})$~\cite{zeng2020pc} to evaluate scanning performance and the path length to evaluate scanning efficiency. 
We generated the ground-truth point cloud $\mathcal{P}_\text{gt}=\{\mathbf{p}_\text{gt}\}$ by applying Poisson Disk Sampling~\cite{yuksel2015sample} to the original mesh model to ensure uniform sampling. 
The point cloud $\mathcal{P}_t$ is generated from $\mathcal{D}_t$ captured at each camera pose.
All $\mathcal{P}_t$ were recorded, and at the end of each inference, voxel downsampling was applied to the accumulated $\mathcal{P}_{1:t}$ to obtain the evaluation point cloud $\mathcal{P}=\{\mathbf{p}\}$. $\mathcal{P}$ was compared with $\mathcal{P}_\text{gt}$ to compute the coverage. 
The coverage $C(\mathcal{P})$ is given as follows:

\begin{equation}
    C(\mathcal{P}) = \frac{1}{{\left| {{\mathcal{P}_\text{gt}}} \right|}}\mathop {\sum }\limits_{\mathbf{p} \in \mathcal{P}} {\mathcal{U}}\left( {\min_{{\mathbf{p}_\text{gt}} \in {\mathcal{P}_\text{gt}}} {\left\| {\mathbf{p} - {\mathbf{p}_\text{gt}}} \right\|}_2} - \epsilon \right),
\end{equation}
where $\mathcal{U}$ is the Heaviside step function, and $\epsilon$ is a distance threshold.

\begin{figure}[t]
  \centering
  \begin{subfigure}{0.45\linewidth}
    \centering
    \includegraphics[width=\linewidth]{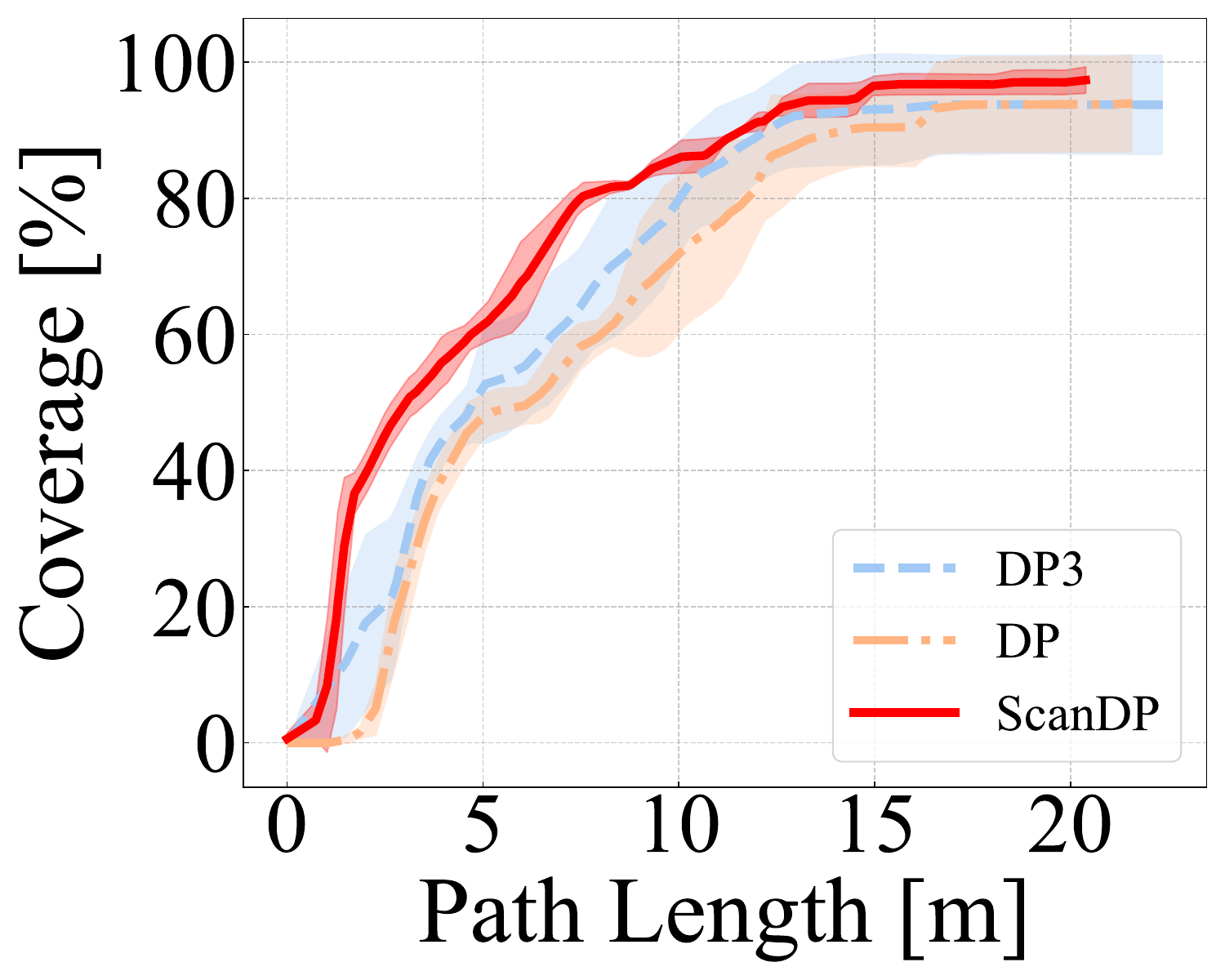}
    \caption{Bunny}
  \end{subfigure}
  \begin{subfigure}{0.45\linewidth}
    \centering
    \includegraphics[width=\linewidth]{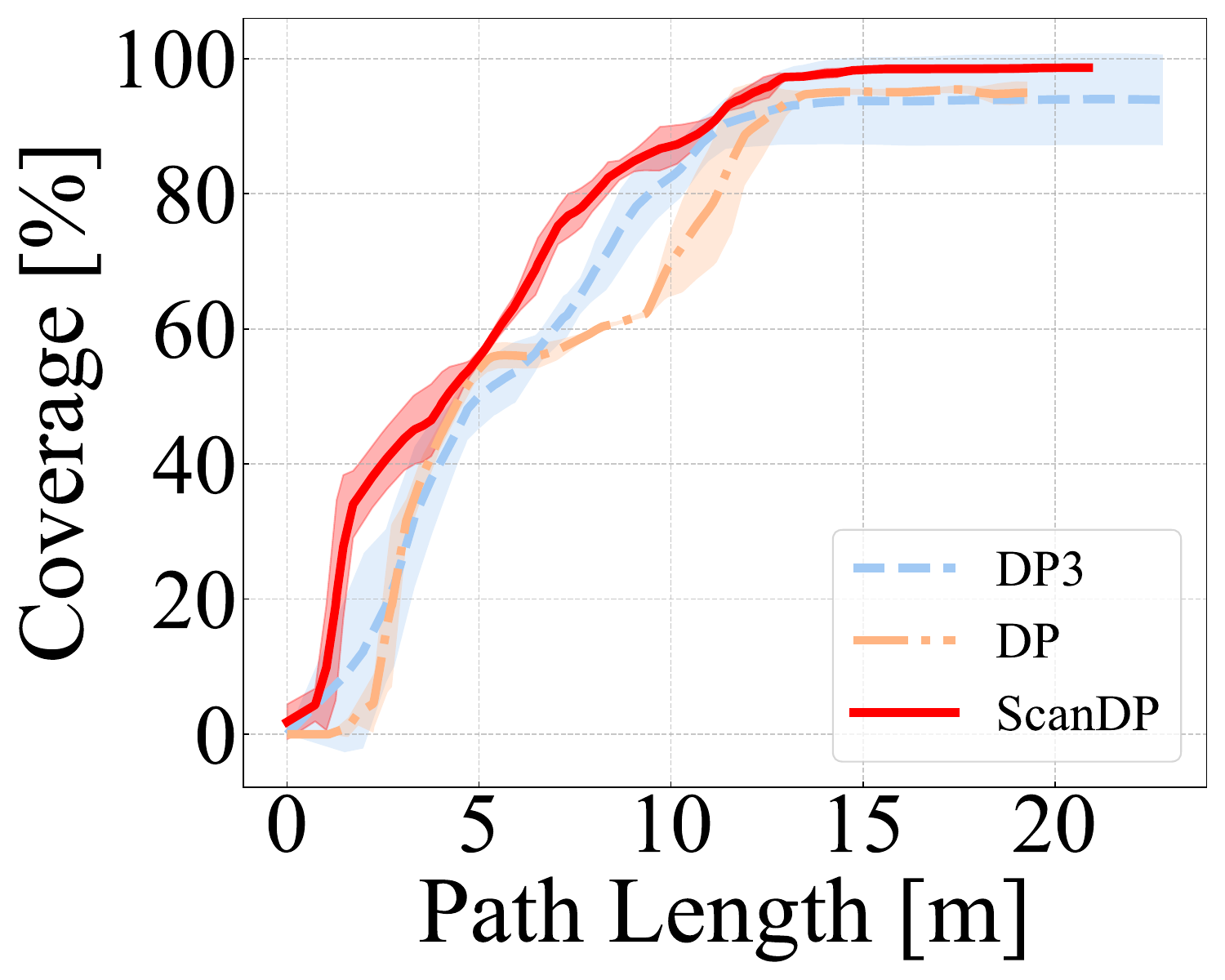}
    \caption{Armadillo}
  \end{subfigure}
  \begin{subfigure}{0.45\linewidth}
    \centering
    \includegraphics[width=\linewidth]{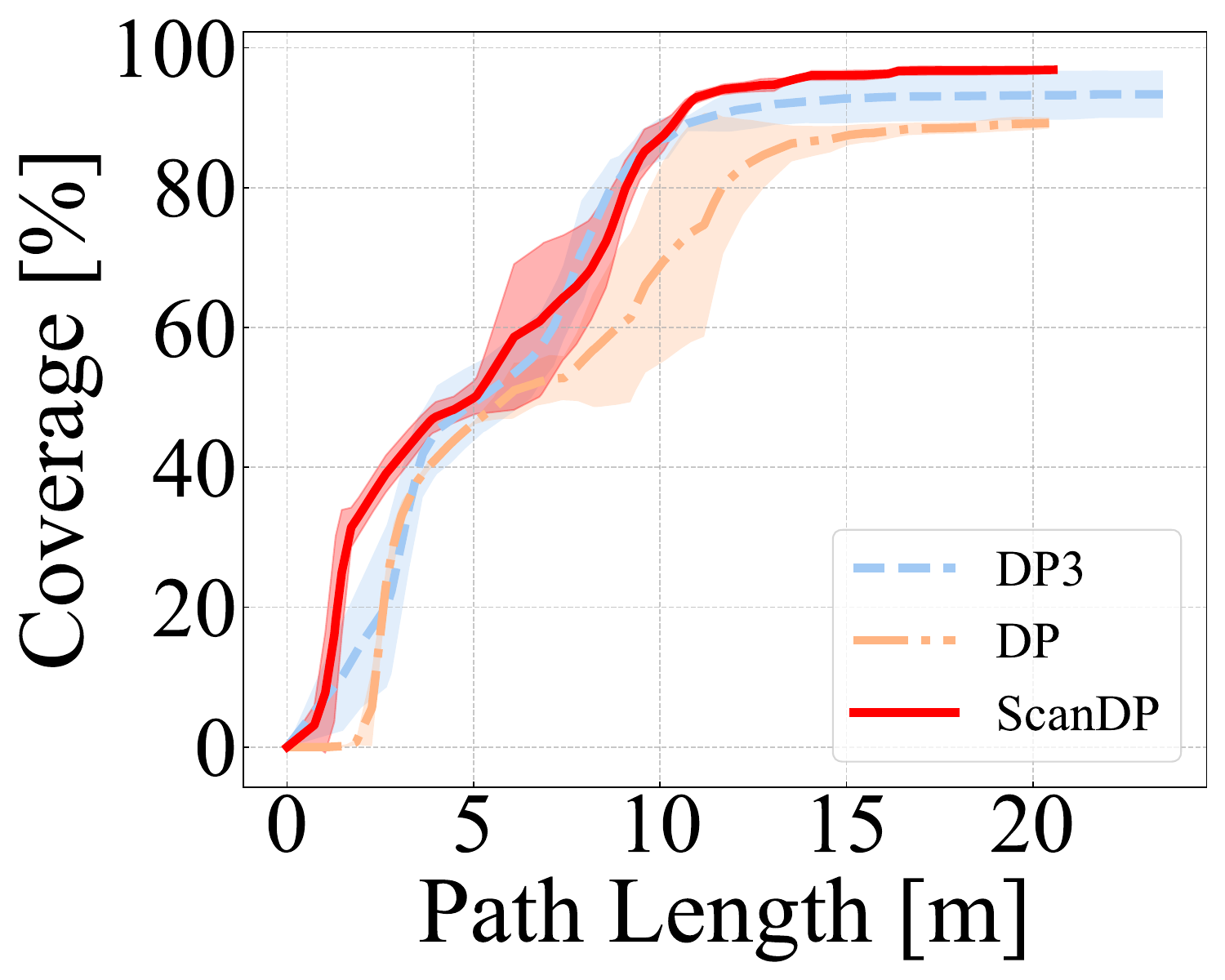}
    \caption{Dragon}
  \end{subfigure}
  \begin{subfigure}{0.45\linewidth}
    \centering
    \includegraphics[width=\linewidth]{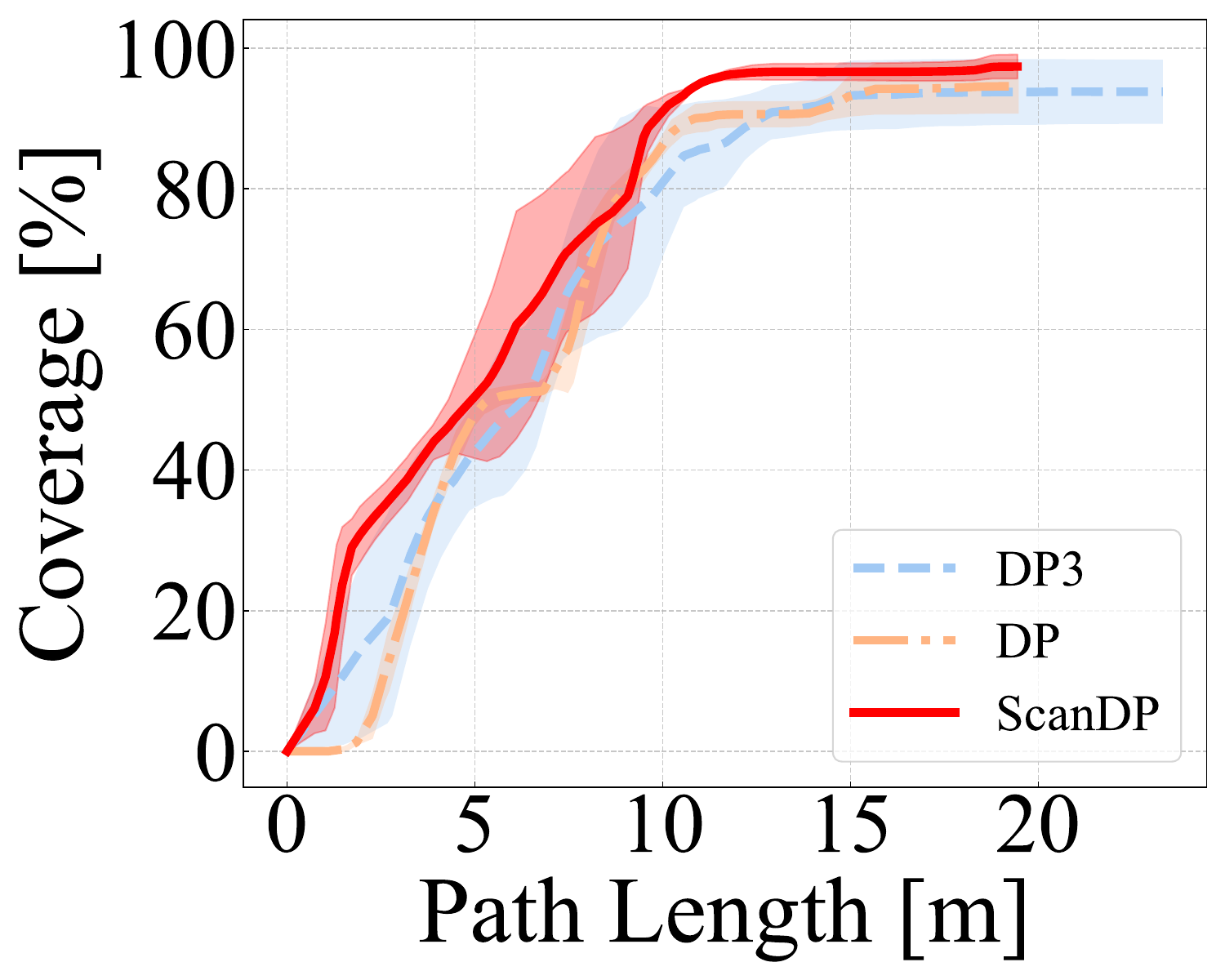}
    \caption{Spot}
  \end{subfigure}
  \caption{Path length comparison (Scale$\times$1.0)}

  \label{fig:x1}
\end{figure}

\begin{figure}[t]
  \centering
  \begin{subfigure}{0.45\linewidth}
    \centering
    \includegraphics[width=\linewidth]{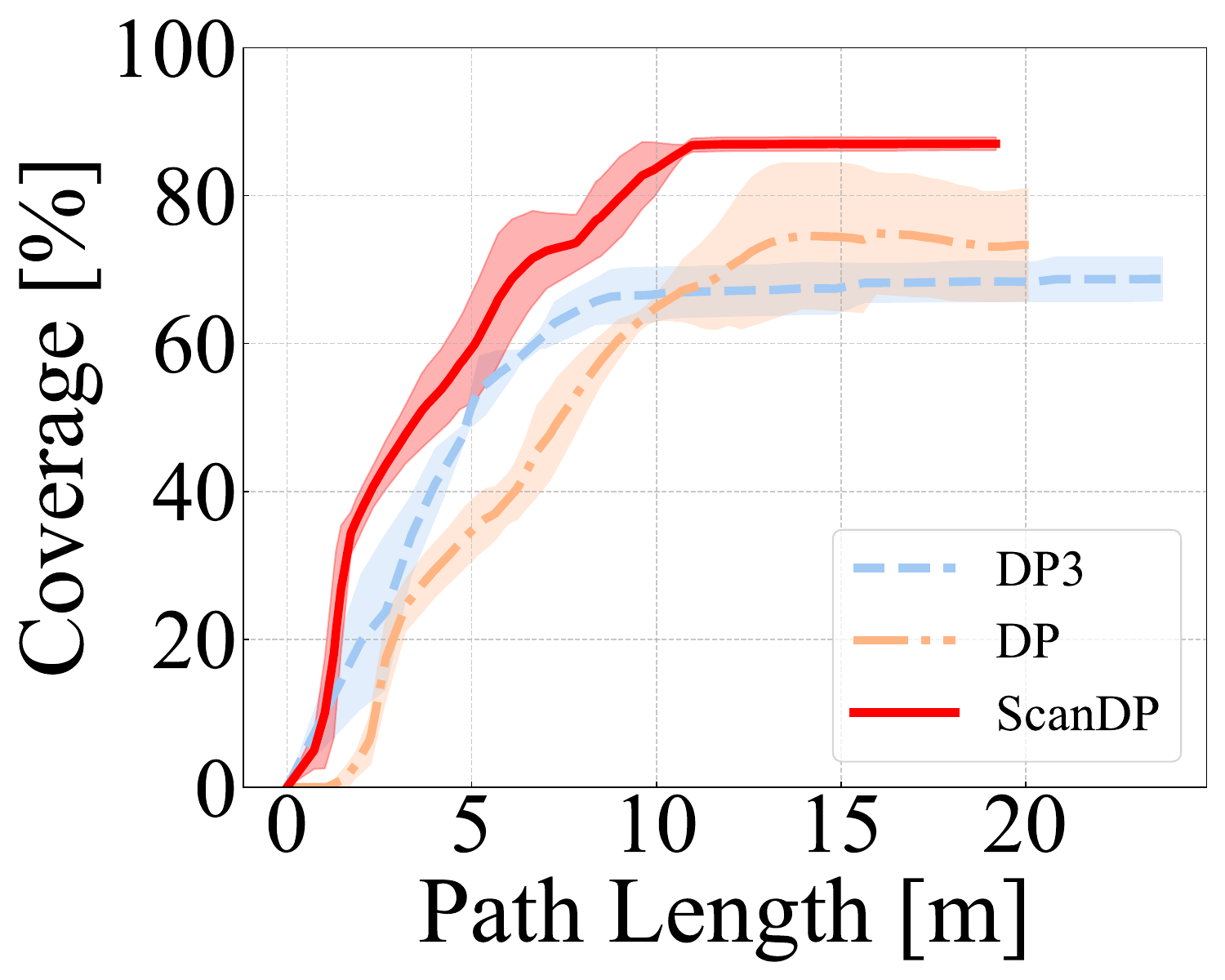}
    \caption{Bunny}
  \end{subfigure}
  \begin{subfigure}{0.45\linewidth}
    \centering
    \includegraphics[width=\linewidth]{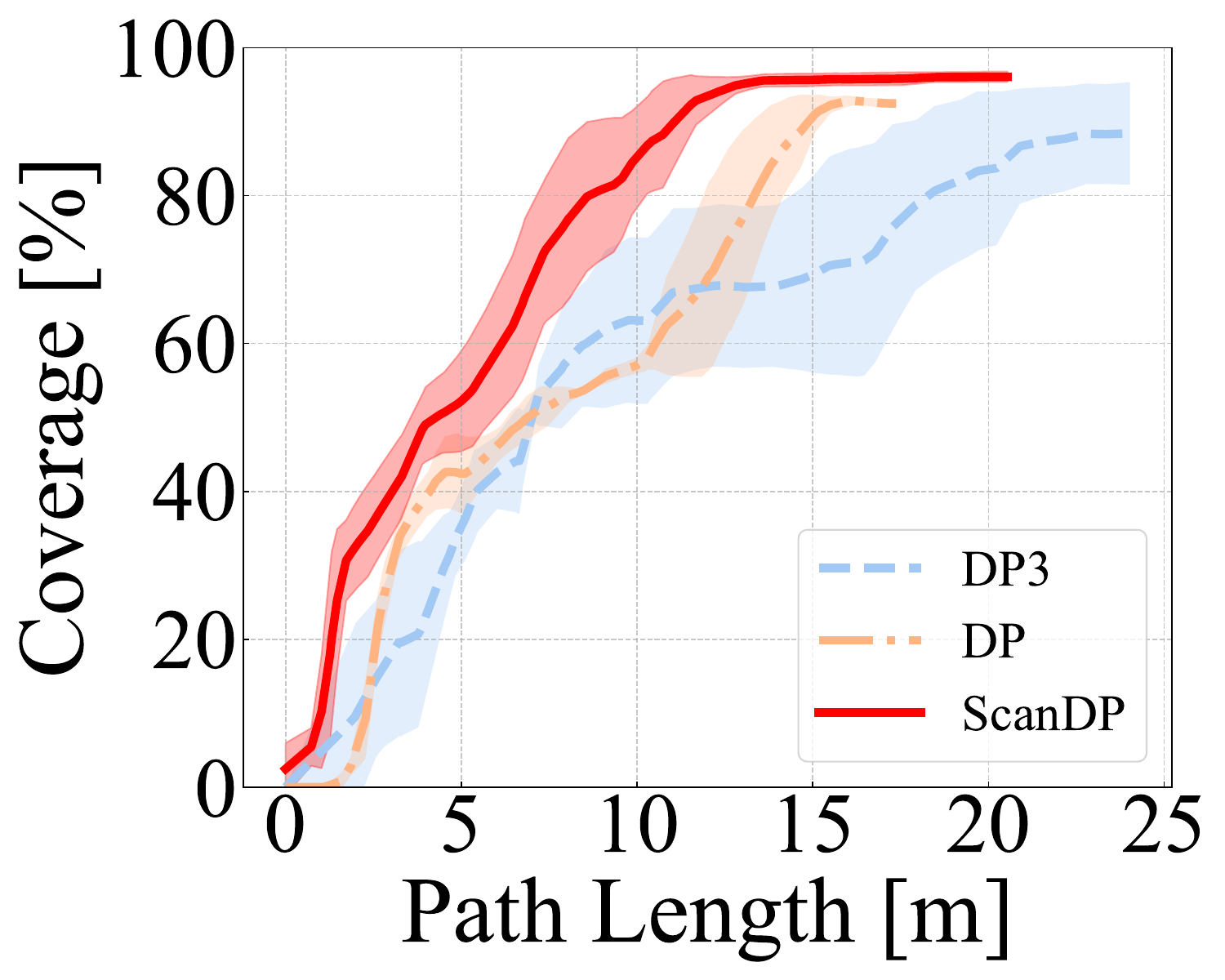}
    \caption{Armadillo}
  \end{subfigure}
  \begin{subfigure}{0.45\linewidth}
    \centering
    \includegraphics[width=\linewidth]{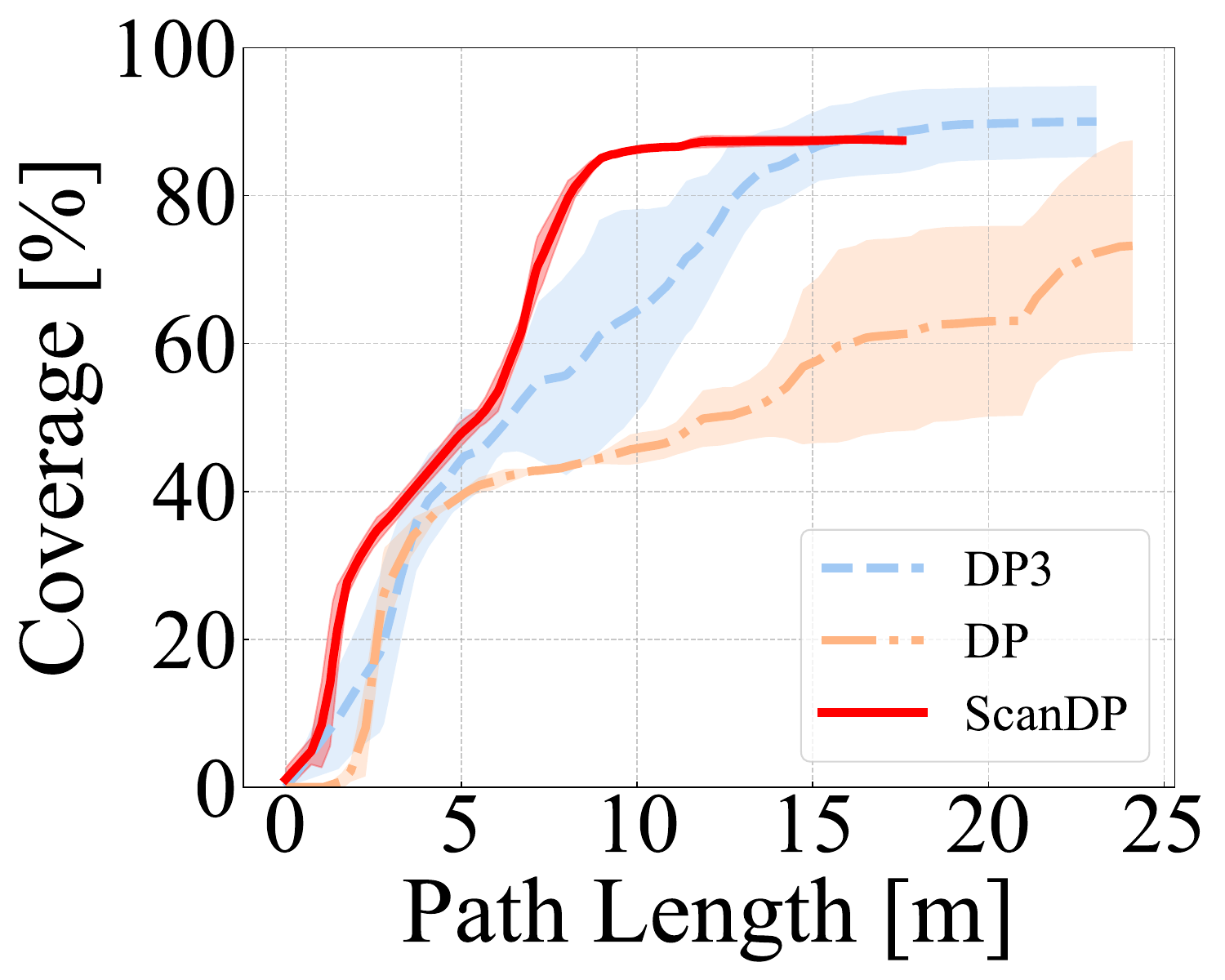}
    \caption{Dragon}
  \end{subfigure}
  \begin{subfigure}{0.45\linewidth}
    \centering
    \includegraphics[width=\linewidth]{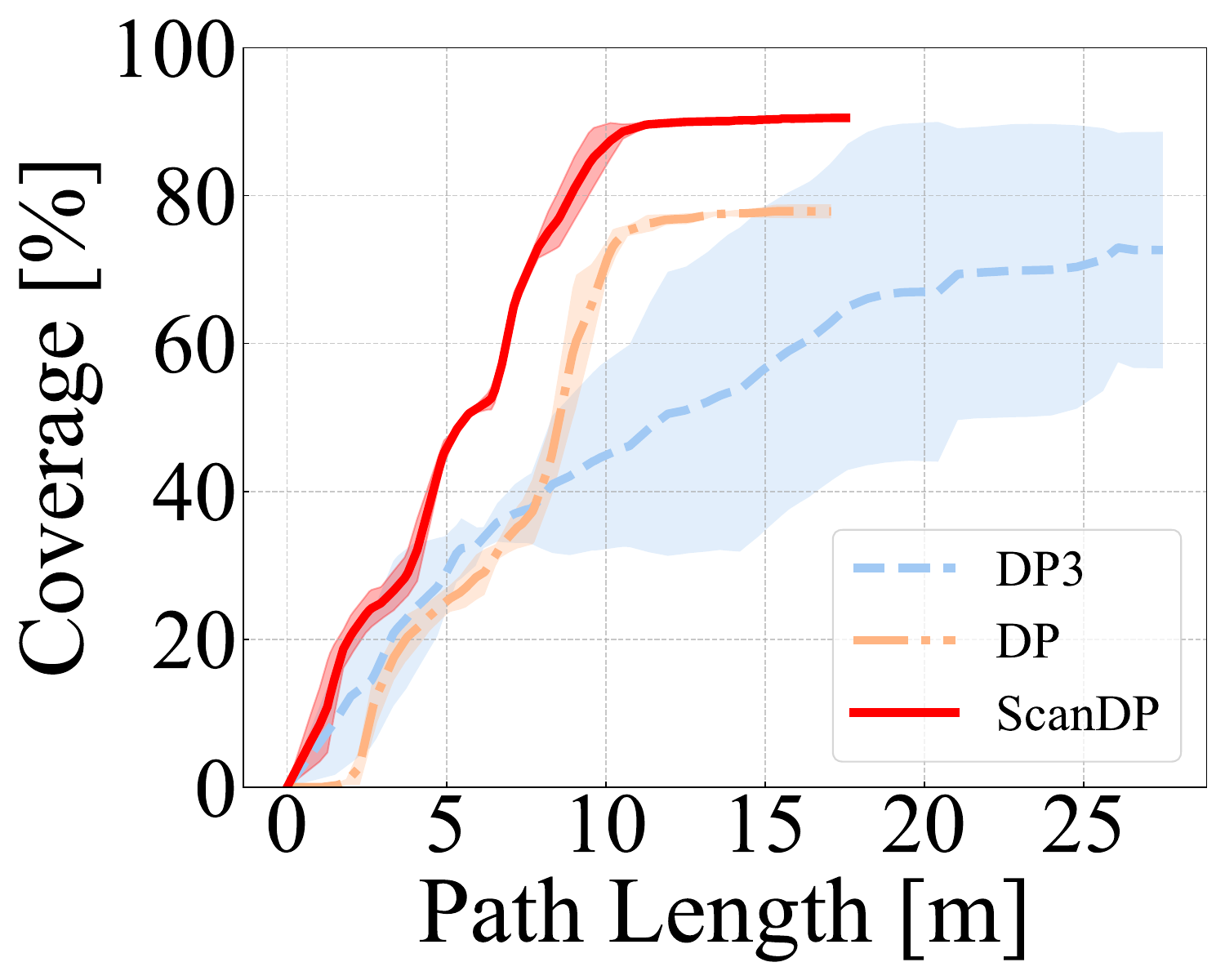}
    \caption{Spot}
  \end{subfigure}
  \caption{Path length comparison (Scale$\times$1.5)}
  \label{fig:x1_5}
\end{figure}

\subsection{Performance Comparison}
We conducted experiments with three different initial positions. 
Each trial ran for a fixed number of 500 steps. 
As shown in Tab.~\ref{tab:coverage}, ScanDP consistently achieved high coverage with shorter paths.
While 3D Diffusion Policy achieved coverage close to that of ScanDP, its performance significantly deteriorated depending on the initial camera position.
Across unseen objects with different sizes, ScanDP achieved 94.0$\pm$4.3\% coverage, while DP and DP3 achieved 87$\pm$11\% and 89$\pm$11\%. 

The main reason is better recovery of self-occluded regions: DP and DP3 often revisit visible areas, while ScanDP selects complementary viewpoints that reveal hidden surfaces.
In all methods, larger objects (Scale$\times$1.5) with simpler shapes, such as the Stanford  Bunny or Spot, tended to result in lower coverage.
This is likely because complex objects have more distinctive features and are easier to recognize, whereas simpler shapes lack such feature points.
In the following sections, we evaluate our method and the baseline methods from two perspectives: target size and noise robustness.

\begin{figure}[t]
  \centering
  \includegraphics[width=\linewidth]{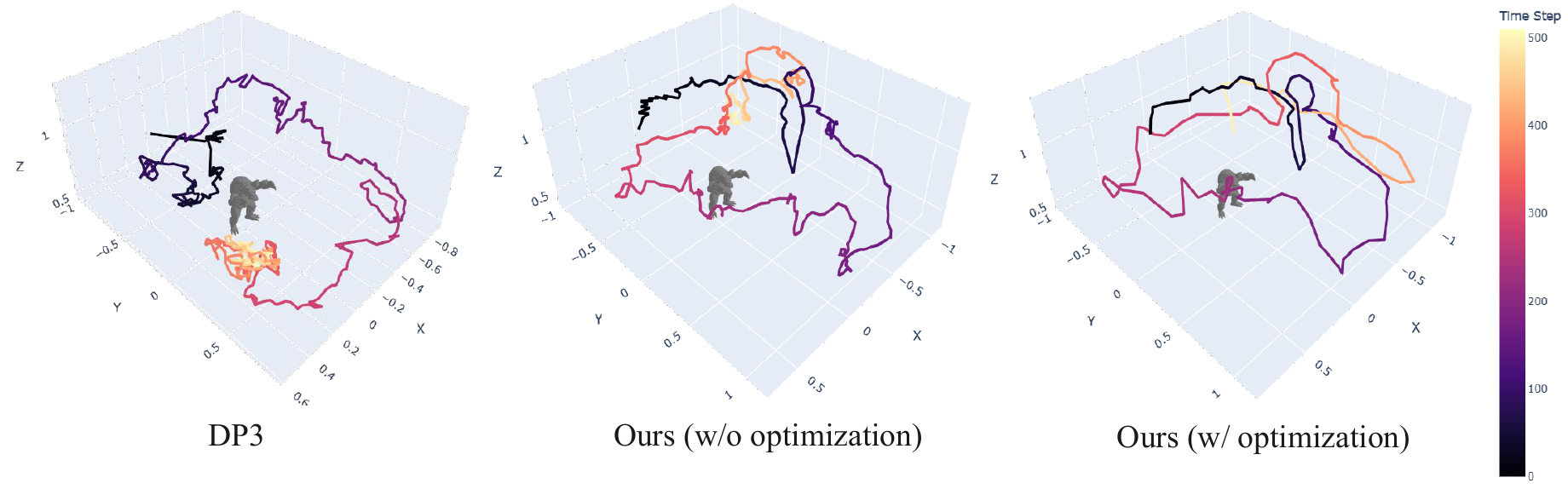}
  \caption{\textbf{Visualization of movement paths from DP3 and ScanDP.} ScanDP with path optimization attains the shortest path length and smoothest movement.
  We find that DP3 tends to get stuck in a particular location when scanning objects not seen during training.}
  \label{fig:path}
\end{figure}

\textbf{Scanning Efficiency.}  
We evaluated the efficiency of camera motion in scanning using the path length.  
Compared to learning-based methods, random and hemisphere approaches result in significantly longer path lengths ($50 \sim 60~m$); therefore, we focused our comparison on DP, DP3 and ScanDP. 
Figures \ref{fig:x1} and \ref{fig:x1_5} show the coverage achieved at different travel distances for object scales of $\times$1.0 and $\times$1.5.  
We show that our method achieves high coverage with a shorter travel distance.  
In contrast, DP3 and our method without path optimization include many redundant motions, leading to inefficiency and longer path lengths. 
With path optimization, ScanDP reduces the total travel distance by an average of 32\% compared to no optimization.
On the other hand, as shown in Fig. \ref{fig:path}, our method with path optimization resulted in the smoothest path.

\textbf{Noise Resistance.}
We evaluated the effect of adding Gaussian noise to the input depth map $\mathcal{D}_t$.
This evaluation simulates real scenarios where RGB-D cameras often capture depth maps with noise, such as when observing objects with reflective or under challenging lighting conditions.
The same objects, initial positions, and sizes as in training were used for consistency in evaluation.  
For comparison, we selected DP3 and ScanDP, as they both take the depth map as input and are thus susceptible to noise.  

From Tab.~\ref{tab:noise_gaussian}, ScanDP maintained 88\% coverage even when the Gaussian noise added to $\mathcal{D}_t$ was 0.1.  
In contrast, DP3 experienced a 20\% drop in coverage when just 0.01 of noise was added.  
We attribute this robustness to the characteristics of the Occupancy Grid Map (OGM).  
OGM assigns independent probability values to each grid cell and integrates multiple observations through Bayesian updates, which averages out the effects of single noisy observations.  
As a result, OGM demonstrates improved robustness against noise.  
Moreover, applying a median filter to the noise-added $\mathcal{D}_t$ before inputting it into the model did not improve the results of DP3.

\begin{table}[t]
    \centering
    \small
    \caption{Coverage [\%] under noisy inputs.}
 \resizebox{\linewidth}{!}{
    \begin{tabular}{lccc}
      \toprule
      std
        & \begin{tabular}{c}
          \includegraphics[width=0.2\linewidth]{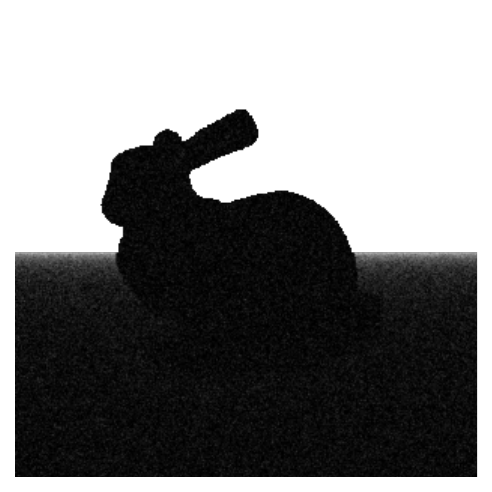} \\ 
          0.01
        \end{tabular}
        & \begin{tabular}{c}
          \includegraphics[width=0.2\linewidth]{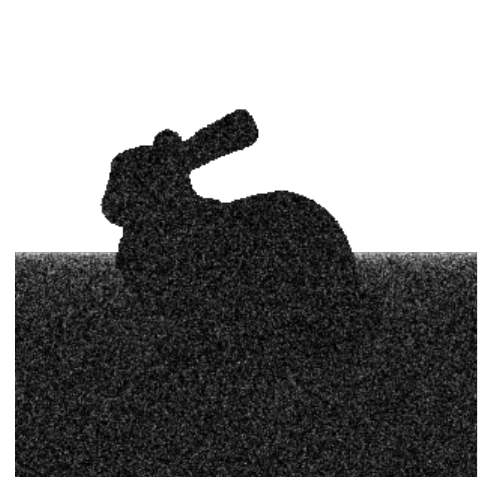} \\ 
          0.1
        \end{tabular}
        & \begin{tabular}{c}
          \includegraphics[width=0.2\linewidth]{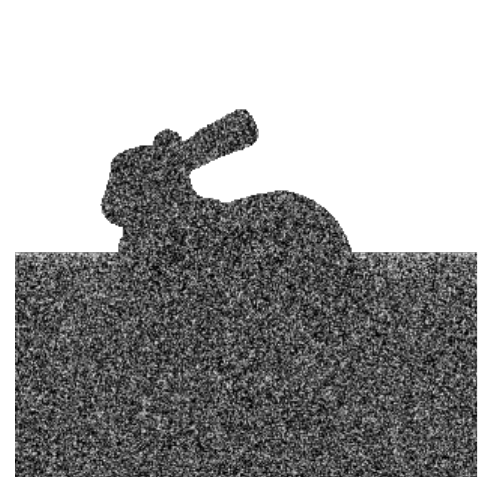} \\ 
          1
        \end{tabular} \\
      \midrule
      DP3\cite{ze20243d} & 73.76  & 74.20  & 69.39 \\
      \textbf{ScanDP}     & \textbf{98.00} & \textbf{88.91} & \textbf{90.44} \\
      \bottomrule
    \end{tabular}
    \label{tab:noise_gaussian}
  }
\end{table}

\textbf{FoV generalization.}
We compared the performance of our method with different field of view (FoV) settings.
We referred to the Intel RealSense L515, D435, and D415 cameras, which have different FoV for experimental settings. 
As shown in Tab.~\ref{tab:coverage_fov}, the construction of an OGM enables stable scanning performance across different FoVs, demonstrating that the method is robust to variations of the FoV.

\begin{table}[t]
  \centering
  \small
  \caption{Coverage [\%] under different FoV.}
  \label{tab:coverage_fov}
  \resizebox{\linewidth}{!}{
  \begin{tabular}{lccc}
    \toprule
    Camera
      & \begin{tabular}{c}
        L515 \\($70^\circ \times 43^\circ$)
      \end{tabular}
      & \begin{tabular}{c}
        D435 \\($87^\circ \times 58^\circ$)
      \end{tabular}
      & \begin{tabular}{c}
        D415 \\($65^\circ \times 40^\circ$)
      \end{tabular} \\
    \midrule
    DP3\cite{ze20243d}  & 58.36  & 60.89  & 61.97 \\
    \textbf{ScanDP}     & \textbf{89.44} & \textbf{83.13} & \textbf{97.40} \\
    \bottomrule
  \end{tabular}
  }
\end{table}

\subsection{Ablations}
We analyze the impact of key design choices in ScanDP using the same evaluation protocol as the main experiments.
We report coverage and path length on unseen objects after the fixed 500-step budget.
Tab.~\ref{tab:ablation} summarizes configuration changes and their outcomes.

We compared the default unthresholded OGM with thresholded OGM (classifying cells into $Occupied$, $Free$, and $Unknown$), and thresholded OGM markedly degraded performance.
This suggests that preserving continuous changes in occupancy probability $p$ is important for policy encoding.

Increasing the grid resolution to 0.01~m within the same spatial range exceeded our available GPU memory and could not be evaluated (N/A in Tab.~\ref{tab:ablation}).
Conversely, lowering the resolution to 0.04~m reduced fidelity for small or self-occluded parts, decreasing coverage to 65.49\%.

\begin{table}[t]
  \centering
  \small
  \caption{\textbf{Ablation results.} 
  The default configuration (probabilistic OGM and 0.02~m grid size) gives the best overall trade-off.
  }
  \resizebox{\linewidth}{!}{
    \begin{tabular}{lcc}
      \toprule
      Variant & Coverage [\%]$\uparrow$ & Path length [m]$\downarrow$ \\
      \midrule
      \textbf{ScanDP} & \textbf{97.84} & \textbf{19.56} \\
      \midrule
      Thresholded OGM & 42.38 & 51.72 \\
      Grid size = 0.01 m & N/A & N/A \\
      Grid size = 0.04 m & 65.49 & 21.58 \\
      \bottomrule
    \end{tabular}
    \label{tab:ablation}
  }
\end{table}

\section{Real-World Experiments}

\begin{figure}[t]
  \centering
  \includegraphics[width=0.85\linewidth]{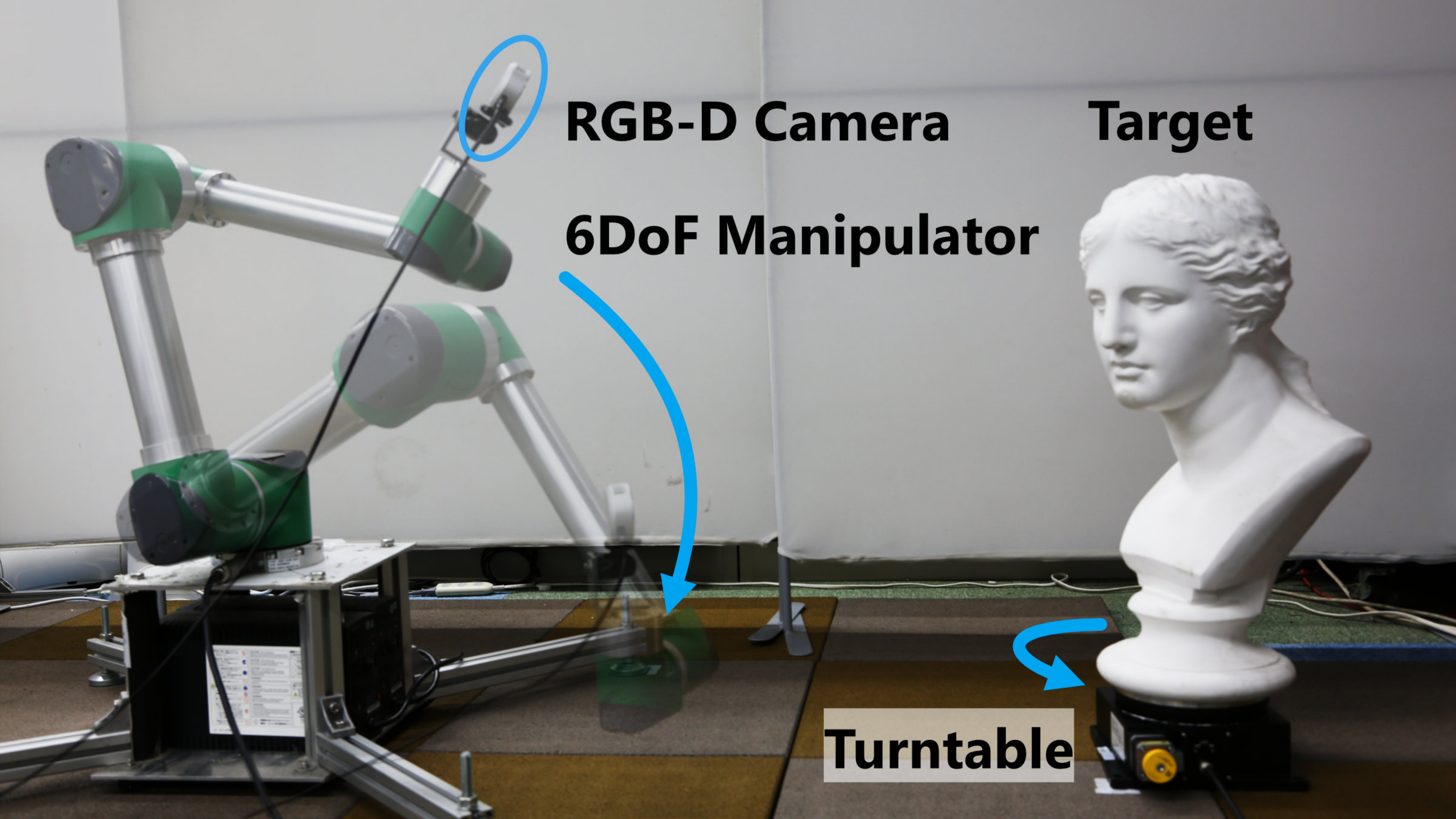}
  \caption{\textbf{Real-world experiment setup.} 
  We use a 6-DoF manipulator with an RGB-D sensor, combined with a turntable to provide an additional rotational degree of freedom. 
  }
  \label{fig:setup_real}
\end{figure}

\subsection{Setup}
\textbf{Real-World Environment.}
All real-world experiments were conducted on ROS Noetic with a 6-DoF manipulator (Nidec NSR1105N-8601), a turntable (SURUGA SEIKI KS402-180C), and an Intel RealSense L515 depth sensor ($240 \times 320$), as shown in Fig.~\ref{fig:setup_real}.
Because the manipulator workspace is limited for full-around object observation, we combined the manipulators motion with turntable rotation to provide an additional rotational degree of freedom while maintaining safe and feasible viewpoints.

\textbf{Dataset.}
We collected five real-world trajectories using an Intel RealSense L515 and RTAB-Map~\cite{labbe2019rtab} for localization, and trained the policy with our proposed method.
For real-world evaluation, we used an unseen object that was not included in the training dataset.
Settings were kept identical to Sec.~\ref{sec:experiments}, including the OGM spatial range, grid resolution, and action/observation horizons.

\textbf{Evaluation Metrics.}
We used the same coverage and path-length definitions as in Sec.~\ref{sec:experiments}.
Ground-truth point clouds were acquired in advance with a handheld scanner (Artec Leo).
For evaluation, background points were removed by SAM3~\cite{carion2025sam}.
For DP3, input point clouds were cropped as in the original paper~\cite{ze20243d}.
Following simulation experiments, each method was evaluated from three different initial camera poses.

\subsection{Performance Comparison}
\begin{figure}[t]
  \centering
  \includegraphics[width=0.45\linewidth]{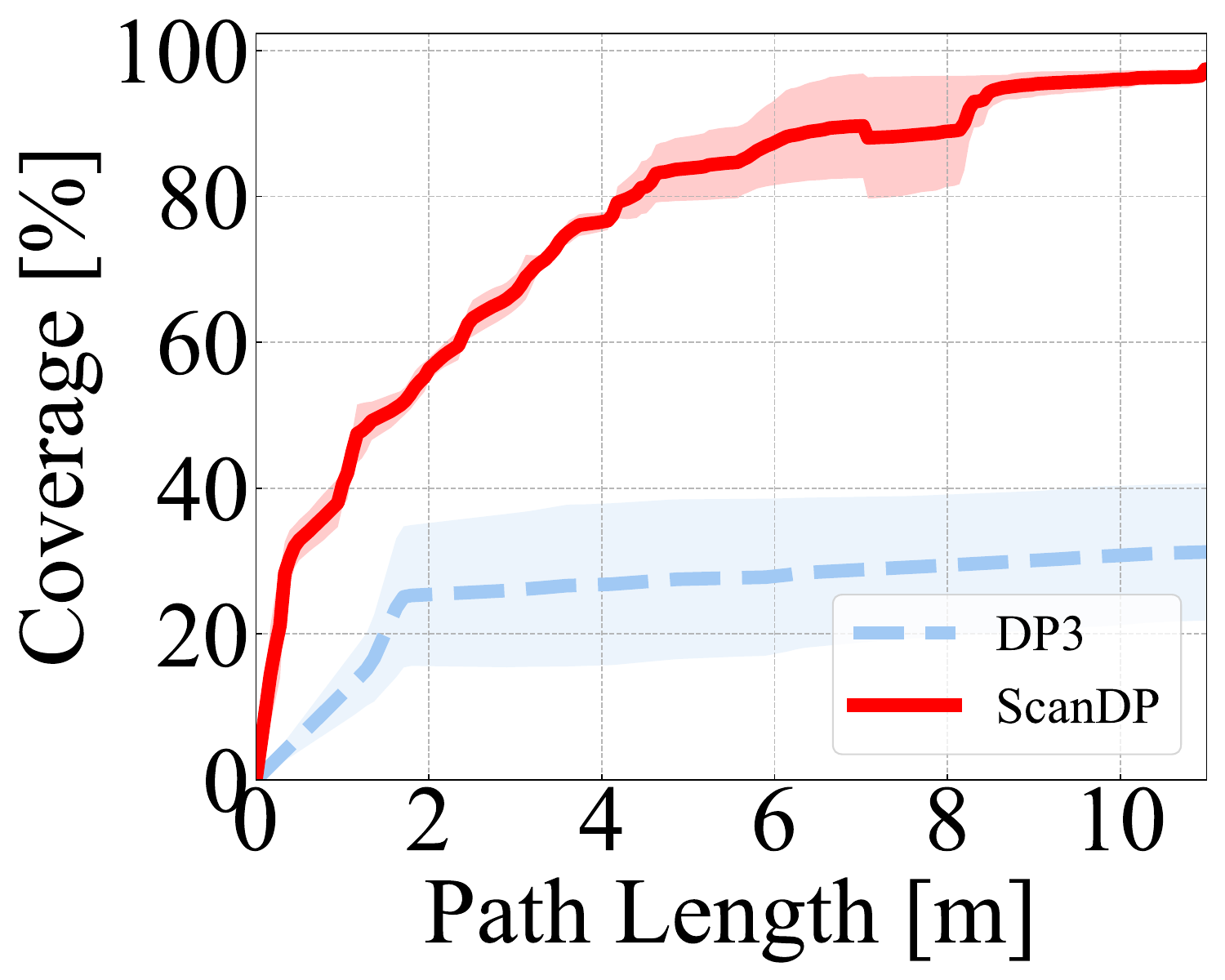}
  \caption{\textbf{Real-world Evaluation Results.} ScanDP demonstrates feasible and stable scanning behavior in real-world conditions.}
  \label{fig:result_real}
\end{figure}

Fig.~\ref{fig:result_real} summarizes the real-world evaluation results.
ScanDP achieved a coverage of 95$\pm$2.0\% across all trials, outperforming DP3's 33$\pm$10.0\%, demonstrating consistent performance with lower variance.
ScanDP integrates observations over time into a global occupancy belief state, whereas DP3 relies mainly on recent point cloud observations.
With the OGM, ScanDP remains robust to transient sensing artifacts and partial occlusions, and continues to select informative next viewpoints even after a temporary loss of target visibility.

\section{Conclusion}
\label{sec:conclusion}
In this work, we proposed ScanDP, an automatic 3D scanning method based on Diffusion Policy.
We evaluated its performance using the progression of coverage and path length over a fixed number of steps.
The results demonstrated that our method achieved superior generalization performance compared to existing works, even with a small amount of training data, and performed well across different initial states, object morphology, and sizes.
Moreover, ScanDP improves safety and efficiency, which were not guaranteed by previous methods.
Under noisy inputs, our method also exhibited significantly greater robustness than prior approaches.

ScanDP leverages the OGM features with sparse convolutions, but this approach may not scale well when extracting features across a larger spatial area while keeping the grid size fixed.
Furthermore, since the training data is based on human motion, domain adaptation is necessary to align the learned policy with robot kinematics.
In future work, we aim to extend our method to handle large-scale scanning and simultaneous scanning of multiple objects.



\section*{ACKNOWLEDGMENT}
This work was funded by JSPS KAKENHI, grant numbers JP24K21173 and JP24H00351. This work was also supported by the UTokyo-SMBC Forest GX project.



\bibliographystyle{ieeetr}
\bibliography{references} 

\end{document}